\documentclass[iicol,pdflatex,sn-mathphys-num]{sn-jnl}

\usepackage{graphicx}%
\usepackage{multirow}%
\usepackage{amsmath,amssymb,amsfonts}%
\usepackage{amsthm}%
\usepackage{mathrsfs}%
\usepackage[title]{appendix}%
\usepackage{xcolor}%
\usepackage{textcomp}%
\usepackage{manyfoot}%
\usepackage{booktabs}%
\usepackage{algorithm}%
\usepackage{algorithmicx}%
\usepackage{algpseudocode}%
\usepackage{listings}%
\usepackage{makecell}
\usepackage{boldline, hhline}
\usepackage{colortbl} 
\usepackage{amsmath}
\usepackage{amssymb}
\usepackage{tabularx}
\usepackage{arydshln}
\usepackage{array}
\usepackage{microtype}

\newcolumntype{C}{>{\centering\arraybackslash}X}

\theoremstyle{thmstyleone}%
\theoremstyle{thmstyletwo}%

\theoremstyle{thmstylethree}%
\definecolor{nice-blue}{HTML}{0071bc}
\definecolor{nice-purple}{HTML}{984EA3}

\begin{document}

\title[Article Title]{Dual-Edged Homogeneous-Modality Similarity: Towards Visible-Infrared Modality-Incomplete Person Re-Identification with Modality Adaptive Matching}

\author[1,2]{\fnm{Xin} \sur{Xu}}\email{xuxin@wust.edu.cn}
\author[1]{\fnm{Shuhao} \sur{Zhan}}\email{1776351000@wust.edu.cn}
\author[1]{\fnm{Wei} \sur{Liu}}\email{liuwei@wust.edu.cn}
\author[3]{\fnm{Zheng} \sur{Wang}}\email{wangzwhu@whu.edu.cn}
\author*[4]{\fnm{Kui} \sur{Jiang}}\email{jiangkui@hit.edu.cn}
\author[5]{\fnm{Chia-Wen} \sur{Lin}}\email{cwlin@ee.nthu.edu.tw}




\affil[1]{\orgdiv{School of Computer Science and Technology}, 
  \orgname{Wuhan University of Science and Technology}, 
  \orgaddress{\city{Wuhan}, \postcode{430065}, \state{Hubei}, \country{China}}}

\affil[2]{\orgdiv{Hubei Province Key Laboratory of Intelligent Information Processing and Real-time Industrial System}, 
  \orgname{Wuhan University of Science and Technology}, 
  \orgaddress{ \city{Wuhan}, \postcode{430081}, \state{Hubei}, \country{China}}}

\affil[3]{\orgdiv{School of Computer Science}, 
  \orgname{Wuhan University}, 
  \orgaddress{\city{Wuhan}, \postcode{430079}, \state{Hubei}, \country{China}}}

\affil*[4]{\orgdiv{Faculty of Computing}, 
  \orgname{Harbin Institute of Technology}, 
  \orgaddress{\city{Harbin}, \postcode{150001}, \state{Heilongjiang}, \country{China}}}

\affil[5]{\orgdiv{Department of Electrical Engineering and the Institute of Communications Engineering}, 
  \orgname{National Tsing Hua University}, 
  \orgaddress{ \city{Hsinchu}, \postcode{30048}, \country{Taiwan}}}

\abstract{Visible-Infrared Person Re-Identification (VI-ReID) conventionally operates under a closed-world assumption, where queries and galleries are from heterogeneous modalities. However, in open-world scenarios, both sets are likely to contain a mixture of homogeneous and heterogeneous modality images. A query may consist of visible-only, infrared-only, or mixed-modality images, while galleries present multi-modal images over long-term collection. Under these conditions, existing VI-ReID methods, built on a heterogeneous-modality retrieval paradigm, suffer from three trustworthiness challenges: matching conflicts due to high homogeneous-modality similarity, interference from modality uncertainty, and robustness degradation induced by unknown modality combinations. Consequently, they fail to meet the requirements of trustworthy visual recognition in reliability, consistency, and dynamic adaptability. To address these challenges, we formalize the Visible-Infrared Modality-Incomplete Re-Identification (VIMI-ReID) task. We reorganize existing datasets to construct the SYSU-VIMI and RegDB-VIMI benchmarks for evaluation. The unpredictable modality combinations and inherent similarity of homogeneous-modality samples in VIMI-ReID cause a significant performance drop in existing VI-ReID methods. To this end, we propose the Modality Adaptive Matching Transformer (MAMT). It employs a Divergence Transformer Module (DTM) and a Shared Transformer Module (STM) to extract modality-specific and modality-shared features, respectively. Guided by a divergence loss, the DTM enriches modality-specific features with modality-style information to enhance discriminability within the same modality. Meanwhile, a Modality Adaptive Matching Module (MAM) dynamically fuses features according to the query-gallery modality relationship, enabling stable matching under arbitrary and uncertain modality conditions. Extensive experiments on the VIMI benchmarks demonstrate the effectiveness and adaptability of MAMT.}




\maketitle

\section{Introduction}\label{sec1} 
Person Re-IDentification (ReID)~\cite{Yin2020Fine, Liu2024Cloth-aware, Ye2025Transformer} aims to retrieve specific individuals across heterogeneous camera networks, playing a vital role in security and surveillance applications. Although visible-spectrum ReID methods have achieved remarkable accuracy, their performance severely degrades in low-light conditions---precisely when criminal activity is most prevalent---due to the inherent limitations of visible-light cameras. The proliferation of multi-modality camera systems has mitigated this issue, as infrared sensors can capture clear pedestrian images even under poor illumination. This advancement has shifted the ReID paradigm from homogeneous-modality to heterogeneous-modality retrieval, where a query captured from one modality is matched against a gallery from another, spurring extensive research on Visible-Infrared Person Re-Identification (VI-ReID)~\cite{Wu2017RGB, Wu2020RGB-IR, Qiu2025Advancing}.

\begin{figure*}[t]
    \centering
    \includegraphics[width=0.94\linewidth]{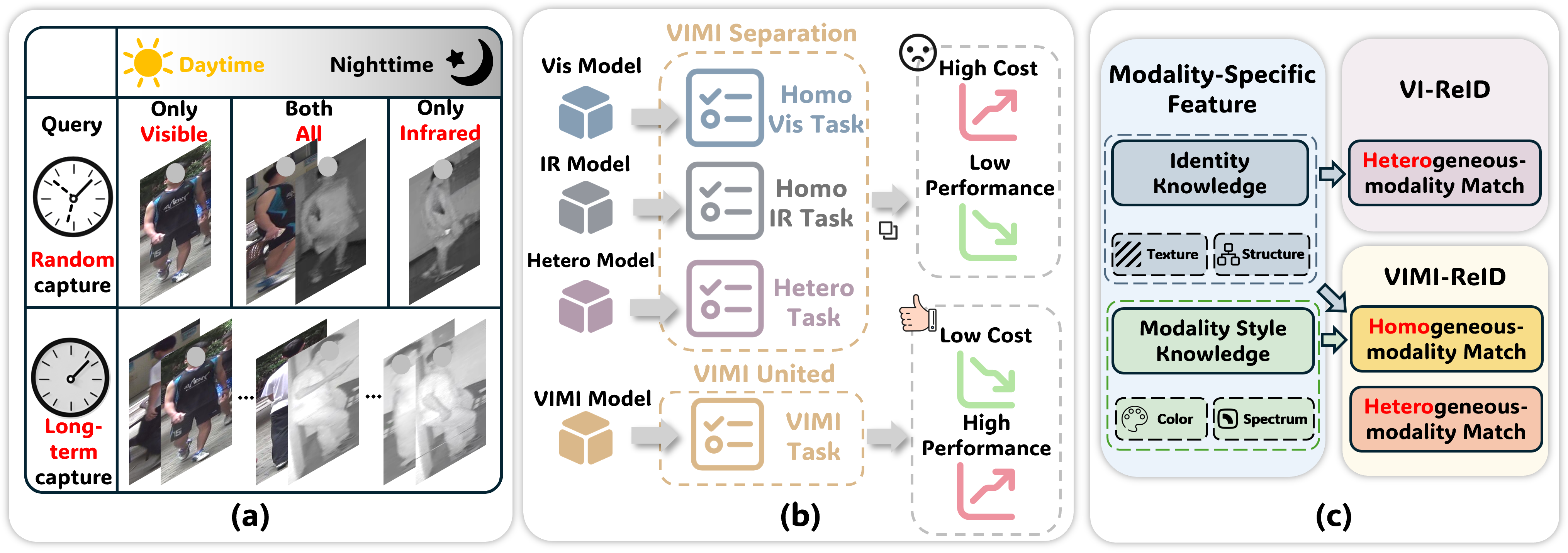}
    \caption{\textbf{(a)} Due to random capture, the query identity may belong to one of three categories: visible-only, infrared-only, or both modalities simultaneously, whereas the gallery set always includes images from both modalities. 
    \textbf{(b)} A comparison is conducted between multi-model and single-model methods in terms of both cost and performance.
    \textbf{(c)} VI-ReID methods typically exploit only the identity-discriminative information in modality-specific features, while ignoring modality-style information. In contrast, our method leverages both identity information and modality-style information, resulting in a more comprehensive feature representation.
    }\vspace{-2mm}
    \label{fig1}
\end{figure*}

Existing VI-ReID methods~\cite{Hu2024Empowering, Pan2024Unified, Cui2024DMA, Sun2024Robust, Zhang2025Modality} are predominantly developed under a closed-world assumption: the query and gallery modalities are fixed and known a priori, being either strictly homogeneous or heterogeneous. While this assumption has facilitated significant progress in heterogeneous-modality retrieval, it substantially diverges from open-world conditions~\cite{Ye2021Channel}, where query modality is uncertain and influenced by environmental factors, time, and device configuration. As illustrated in Figure~\ref{fig1}(a), real-world queries may be visible-only, infrared-only, or mixed-modality, while galleries accumulated over time are inherently multi-modal.

To bridge this gap between closed-world ReID and open-world person retrieval, we propose a new task named Visible-Infrared Modality-Incomplete Person Re-IDentification (VIMI-ReID). The core objective of VIMI-ReID is to perform robust retrieval in a modality-incomplete environment, where queries can be of any modality composition and galleries contain both modalities. This task more accurately reflects the unpredictability of real-world deployments. 

To support this task, we reorganize existing VI-ReID test sets to construct two benchmarks: SYSU-VIMI and RegDB-VIMI. Given the data structure of VIMI-ReID, we categorize matching into homogeneous-modality (visible-to-visible, infrared-to-infrared) and heterogeneous-modality (visible-to-infrared, infrared-to-visible) retrieval. A naive approach is multi-model fusion, where separate models are trained for each modality and used jointly during inference. However, as shown in Figure~\ref{fig1}(b), this strategy incurs high training and deployment costs with limited performance gains. More critically, when applying state-of-the-art VI-ReID methods to VIMI-ReID, we observe a significant drop in mAP despite a slight increase in Rank-1 accuracy.

This phenomenon underscores a critical challenge: the dual role of modality similarity in open-world environments. While beneficial for homogeneous-modality retrieval, this similarity introduces modality conflicts that hinder heterogeneous-modality retrieval. Specifically, it causes samples of different identities from the same modality to be ranked higher than the correct match from a different modality, leading to homogeneous-modality interference. Furthermore, existing VI-ReID methods primarily focus on learning modality-invariant representations, often discarding discriminative modality-style information embedded in modality-specific features. As shown in Figure~\ref{fig1}(c), while this information has marginal utility in traditional VI-ReID, it becomes crucial for discriminability in the VIMI-ReID task.

To address these challenges, we propose the Modality Adaptive Matching Transformer (MAMT), which dynamically fuses modality-shared and modality-specific features by adapting the matching strategy to the query-gallery modality relationship. Our framework consists of a Shared Transformer Module (STM) that learns modality-invariant representations from mixed-modality inputs, and a Divergence Transformer Module (DTM) that enhances modality-specific features from single-modality inputs under the guidance of a divergence loss. A Modality Adaptive Matching Module (MAM) then adaptively fuses these features, enabling robust and conflict-aware matching across any modality combination. 

The main contributions of this work are as follows:
\begin{itemize}
\item \textbf{Task Contribution.}
We identify the limitations of conventional VI-ReID in open-world conditions and propose a more practical and generalizable task, VIMI-ReID. To foster research in this direction, we reorganize existing benchmarks into two new datasets---SYSU-VIMI and RegDB-VIMI.
\item \textbf{Methodological Contribution.}
We diagnose the dual role of modality similarity in VIMI-ReID, which facilitates homogeneous-modality matching but impedes heterogeneous-modality matching. Based on this insight, we design MAMT to address this issue.
\item \textbf{Empirical Contribution.}
Extensive experiments show that MAMT achieves robust and conflict-aware matching under open-world modality-incomplete scenarios, outperforming existing methods on both SYSU-VIMI and RegDB-VIMI.
\end{itemize}

\section{Related Work}
\subsection{Visible Person Re-IDentification}
Visible Person Re-IDentification (V-ReID) aims to retrieve person images across a network of non-overlapping cameras using visible images. 

Recent advancements in deep learning have significantly improved V-ReID performance, especially in addressing challenges related to variations in pose, viewpoint, and occlusion~\cite{Dou2023Hum,ye2024dynamic, Yin2020Fine}. However, limited attention has been given to the degradation of image quality under low-light conditions~\cite{Min2018Bli}. Several studies have explored ReID in low-light scenarios, focusing on image enhancement and learning light-independent features to boost retrieval performance~\cite{Huang2019Illumination, Lu2024Illumination}. Despite these efforts, none have effectively addressed the inherent limitations of RGB cameras in such conditions. These limitations include the inability to capture sufficient detail in dark environments, leading to person identity loss, which remains a significant challenge for V-ReID under low-light conditions. With the increasing prevalence of dual-modal cameras, Wu~\cite{Wu2017RGB} introduced the VI-ReID task to address the limitations of RGB cameras by leveraging a single-stream neural network. This approach provides a robust solution for handling low-light conditions, as infrared images capture complementary information when light is insufficient, improving Re-ID accuracy across varying lighting environments.

\subsection{Visible-Infrared Person Re-IDentification}
Visible-Infrared Person Re-IDentification (VI-ReID) aims to retrieve images from one modality using images from another modality. Research on VI-ReID mainly reduces modality discrepancies through two types of approaches: directly learning more discriminative modality-shared features, and enhancing these shared features with modality-specific representations.

Research on directly learning modality-shared features can be further divided into feature-level and image-level approaches. Feature-level methods mitigate modality discrepancies by applying constraints directly in the feature space, while image-level methods primarily reduce such discrepancies by generating intermediate-modality images. HCL~\cite{Zhu2020Hetero} and HCTL~\cite{Liu2021Parameter} align different modality feature distributions using heterogeneous center constraints, while MMG~\cite{Zhang2021Towards} introduces intermediate modality feature learning to further mitigate modality differences. HOS-Net~\cite{Qiu2024High} enhances feature discriminability through high-order structure modeling and a modality-range identity center contrastive loss. ProtoHPE~\cite{Zhang2023ProtoHPE} refines feature details using a high-frequency compensation strategy, whereas CA~\cite{Ye2024Channel} improves the robustness of modality-shared features via channel enhancement. UNet~\cite{Pan2024Unified} reduces modality discrepancies through conditional image generation. Recent studies~\cite{Hu2024Empowering} have shifted toward large-model paradigms, leveraging textual descriptions to enrich the representation capability of the infrared modality. These methods primarily focus on learning more discriminative modality-shared features to facilitate heterogeneous-modality matching. However, they overlook the fact that such highly discriminative shared features may introduce interference in homogeneous-modality matching under mixed-modality scenarios.

Modality-specific features leverage identity-related information inherent to each modality to compensate for the limitations of shared features, thereby reducing modality discrepancies. Early methods, such as MSN~\cite{Feng2020Learning}, introduced modality-specific representation learning and optimized intra-modality discriminability using modality-specific contrastive loss. Building on this, cm-SSFT~\cite{Lu2020Cross} proposed a shared-specific feature transfer mechanism to enhance heterogeneous-modality matching. FMCNet~\cite{Zhang2022FMCNet} further introduced a feature-level modality compensation strategy to reduce modality discrepancies, while TSME~\cite{Liu2022Revisiting} refined modality-specific compensation techniques to improve modality representation learning. Other approaches focus on efficient utilization of modality-specific features. PIC~\cite{Zheng2022Visible} employs a partial interaction mechanism through collaborative learning to enhance the effectiveness of modality-specific features. MSMNet~\cite{Li2022Visible} utilizes a modality-specific memory network to store and retrieve modality-specific features, improving matching robustness. MUN~\cite{Yu2023Modality} constructs an auxiliary modality and integrates heterogeneous-modality and homogeneous-modality learners to dynamically model both modality-specific and modality-shared representations, mitigating variations in both heterogeneous-modality and homogeneous-modality scenarios. The latest method, IDKL~\cite{Ren2024Implicit}, introduces an implicit discriminative knowledge learning approach, reinforcing modality-shared features by incorporating implicit discriminative signals during feature learning. 
Despite their differences, these methods share a common design philosophy: the specific branch is assigned the task of learning identity-discriminative information to compensate for what is lost during heterogeneous-modality alignment, and is used exclusively for heterogeneous-modality retrieval. This design is effective under the strictly heterogeneous setting of conventional VI-ReID, but becomes inadequate in VIMI-ReID, as such a design discards modality-style information that plays a discriminative role in homogeneous-modality matching.
\section{VIMI-ReID}
\label{3.1}
In this section, we define the VIMI-ReID task and analyze the challenges it faces. 

\subsection{Task and Formulation}

In open-world surveillance systems, visible and infrared cameras are typically deployed simultaneously, and a person may be captured by one or both modalities depending on the time of day and environmental conditions. However, existing ReID paradigms assume that the query and gallery are restricted to a single modality or a fixed modality combination, which limits their applicability to such open-world scenarios. Conventional VI-ReID assumes a fixed retrieval direction, where the query is drawn from one modality and the gallery from the other, forming a strictly heterogeneous setting. V-ReID operates under a homogeneous setting, where all gallery samples share the same modality. Neither paradigm accounts for a more realistic scenario in which the gallery simultaneously contains samples from both modalities and the modality of the query is uncertain, a setting that more accurately reflects the uncertainty of open-world environments.

To address this, we propose Visible-Infrared Modality-Incomplete Person Re-Identification (VIMI-ReID). This task formulates ReID as a mixed-modality retrieval task, thereby realistically reflecting the modality uncertainty in open-world environments. Specifically, the gallery in VIMI-ReID contains identities with visible and infrared images, while the modality of the query is uncertain. A comparison between VIMI-ReID and existing ReID paradigms is summarized in Table~\ref{tab:paradigm_comparison}.

\begin{table}[th]
\caption{Comparison of different ReID paradigms in terms of query modality, gallery modality, and core challenge.}
\label{tab:paradigm_comparison}
\centering
\begin{tabular*}{\columnwidth}{@{\extracolsep{\fill}}lccc}
\toprule
Paradigm & Query & Gallery & Core Challenge \\
\midrule
V-ReID & Vis & Vis & \makecell{Identity\\discrimination} \\ \midrule
\multirow{2}{*}{VI-ReID} & Vis & IR & \multirow{2}{*}{Modality gap} \\
 & IR & Vis & \\
 \midrule
\textbf{VIMI-ReID} & Vis / IR & Vis + IR & \textbf{\makecell{Homogeneous\\-modality\\interference}} \\
\bottomrule
\end{tabular*}\vspace*{-4mm}
\end{table}

A straightforward solution to the VIMI-ReID task is to combine V-ReID and VI-ReID models to handle different modality retrieval scenarios. However, this approach requires training multiple independent models, which not only significantly increases the training cost but also adds to the deployment complexity. In addition, existing studies on single-modality infrared ReID remain limited. Therefore, it is necessary to develop a unified ReID framework that can efficiently address the VIMI-ReID task within a single model.

We now formally define the VIMI-ReID task.
Similar to VI-ReID, the VIMI-ReID task divides the test set into two parts: a query set $\mathcal{Q} = \{q_i, i \in N_q\}$ and a gallery set $\mathcal{G} = \{g_j, j \in N_g\}$, where $N_q$ and $N_g$ represent the total number of samples in $\mathcal{Q}$ and $\mathcal{G}$, respectively. However, unlike VI-ReID, where the query and gallery sets belong to different modalities, the VIMI-ReID test set includes both visible and infrared images in both sets. Specifically, $1/3$ of the identities in the $\mathcal{Q}$ have only visible images, $1/3$ of the identities have only infrared images, and $1/3$ of the identities have both visible and infrared images, while the gallery set $\mathcal{G}$ contains visible and infrared images according to the specific evaluation setting.

\subsection{Task Challenge}


VIMI-ReID inherits the fundamental challenges of both V-ReID and VI-ReID, including accurate identity matching and the large modality gap that hinders feature alignment across modalities. Beyond these inherited challenges, the mixed-modality gallery structure of VIMI-ReID introduces a new and non-trivial issue that does not exist in V-ReID or VI-ReID, namely homogeneous-modality interference. Since the similarity between homogeneous-modality samples is jointly influenced by identity and modality similarity while the similarity between heterogeneous-modality samples relies primarily on identity similarity, homogeneous-modality samples receive an additional similarity boost that causes wrong homogeneous-modality samples to be ranked ahead of correct heterogeneous-modality matches. Unlike conventional ranking errors caused by insufficient identity discriminability, this interference arises even with discriminative features, as it stems from modality similarity rather than insufficient identity discriminability. Thus, it cannot be resolved by improving identity discrimination alone, and constitutes a unique challenge of VIMI-ReID.
\begin{figure*}[t!]
    \centering
    \includegraphics[width=1\linewidth]{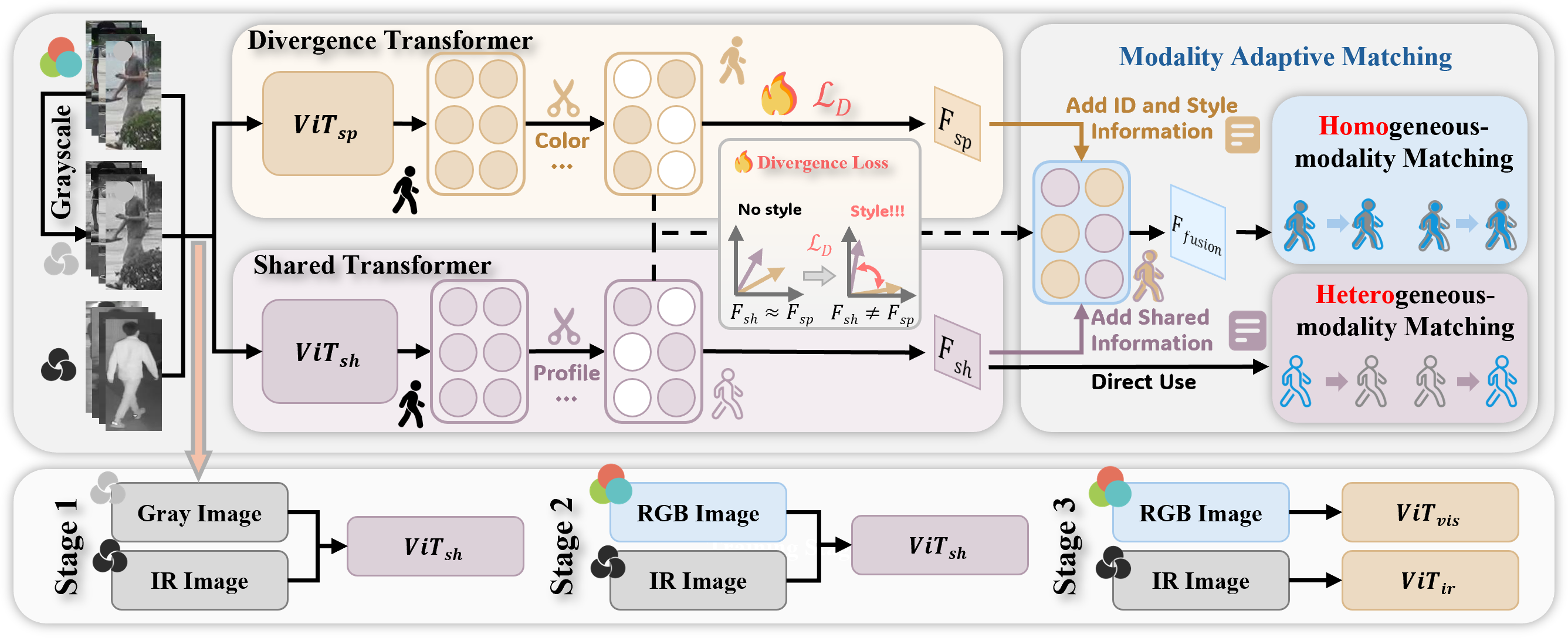}\vspace*{-4mm}
    \caption{MAMT employs a staged training strategy to fully learn both modality-shared features and modality-specific features. Specifically:
    Stage 1: Infrared images and grayscale-converted visible images are input to $ViT_{sh}$ to reduce the modality gap in shared features; Stage 2: RGB images and infrared images are input to $ViT_{sh}$ to further learn discriminative modality-shared representations; Stage 3: RGB and infrared images are separately input to their respective $ViT_{sp}$ branches, and modality-specific features are learned under the $L_D$ loss constraint. In addition, guided by modality labels, MAM activates the modality-specific features only for homogeneous-modality matching, where modality-style information can help distinguish homogeneous-modality samples.
    }\vspace*{-2mm}
    \label{framework}
\end{figure*}

\section{Methodology}
In this section, we introduce MAMT, which is designed for the VIMI-ReID task. MAMT consists of three modules: STM, DTM, and MAM, as shown in Figure~\ref{framework}.
STM serves as the foundation of the framework, learning modality-shared identity features through a progressive training strategy to ensure heterogeneous-modality consistency. 
DTM is built upon the stable shared representations produced by STM: it is introduced only after STM has converged, taking the shared features as a reference signal to learn modality-style information that is ignored by the shared representations through the Divergence Loss. MAM relies on both STM and DTM to perform adaptive matching.


\subsection{Shared Transformer Module}
\label{3.2}
The key to heterogeneous-modality matching lies in learning modality-invariant shared features, allowing the model to effectively extract identity information across different modalities. To achieve this, we propose the \textbf{S}hared \textbf{T}ransformer \textbf{M}odule (STM), which leverages a shared Vision Transformer to extract modality-shared features.

To better capture modality-shared features, a progressive learning approach is adopted to learn modality-shared features~\cite{Lu2023Learning}. In the first stage, the visible image $x_{vis}$ is converted into a grayscale image $x_{gray}$ to learn the fundamental shared features between $x_{gray}$ and infrared images $x_{ir}$. 
This grayscale conversion removes modality-specific color information from visible images~\cite{Zhong2022Grayscale}, thereby reducing the modality gap and encouraging the model to focus on modality-invariant structural cues during early training.
In the second stage, modality-shared features between $x_{vis}$ and $x_{ir}$ are learned.
Based on this motivation, given a pair of input images $(x_{vis},x_{ir})$, the visible image $x_{vis}$ is converted into a grayscale image $x_{gray}$ through a weighted summation, as follows:
\begin{equation}
  x_{gray}= 0.2928x_{vis}^{R}+0.5870x_{vis}^{G}+0.1140x_{vis}^{B},
   \label{eq1}
\end{equation}
where $x_{vis}^{R},x_{vis}^{G},x_{vis}^{B}$ represent the red channel, green channel, and blue channel of the visible image, respectively. 
After processing the three types of images through the shared ViT $ViT_{sh}$, the corresponding image features are extracted. The specific implementation is as follows:
\begin{equation}
    F_{sh}^{vis},F_{sh}^{gray},F_{sh}^{ir}=ViT_{sh}(x_{vis},x_{gray},x_{ir}).
    \label{eq2}
\end{equation}

To enhance the identity discrimination capability of modality-shared features, ID loss $L_{{id}-{sh}}$ and Triplet loss $L_{tri}$ are employed. $L_{{id}-{sh}}$ ensures that features are well-clustered based on identity labels, while $L_{tri}$ refines feature separability by enforcing a margin between positive and negative sample pairs. $L_{tri}$ is formulated as follows:
\begin{equation}
\begin{aligned}
 L_{tri}(X)={\sum_{i=1}^P \sum_{a=1}^K}[m+\max _{p=1 \ldots K}d(x_a^i,x_p^i) \\-
 \min _{\substack{j=1 \ldots P \\
n=1 \ldots K \\
j \neq i}}d(x_a^i,x_n^j)]_{+}
\end{aligned}
,
\label{eq3}
\end{equation}
where $x_a^i$ denotes the feature of the $a$-th sample from the $i$-th identity, $x_p^i$ denotes a positive sample belonging to the same identity as $x_a^i$, and $x_n^j$ denotes a negative sample from a different identity ($j\neq i$). $d(\cdot,\cdot)$ represents the Euclidean distance, $[\cdot]_+$ denotes the hinge loss, and $m$ is the margin parameter that controls the separation between positive and negative sample distances.

To adapt to stage-wise training, we redefine the existing Triplet loss so that it can simultaneously mitigate both intra-modality and inter-modality differences during the training process. The specific calculation is as follows:
\begin{equation}
    L_{tri-sh} = \left\{ 
    \begin{array}{ll}
        L_{tri}(F_{sh}^{gray})+L_{tri}(F_{sh}^{ir}) & Stage\ 1 \\
        L_{tri}(F_{sh}^{vis},F_{sh}^{ir}) & Stage\ 2
    \end{array}
    \right.
    .
    \label{eq4}
\end{equation}
Here, $L_{tri}(F_{sh}^{vis}, F_{sh}^{ir})$ represents the constraint applied to the mixed modality features of both visible and infrared modalities. Stage 1/2 indicates that different Triplet loss functions are used at different training stages to regulate the learning of modality-shared features.

The loss function for the modality-shared features in STM is formulated as follows:
\begin{equation}
    L_{STM}=L_{id\text{-}sh}+L_{tri\text{-}sh}.
    \label{eq5}
\end{equation}

\subsection{Divergence Transformer Module}
\label{3.3}
Modality-shared features are learned during the 1st and 2nd training stages. However, relying solely on shared features can lead to homogeneous-modality interference, because these features mainly focus on modality-invariant information and ignore modality-specific information, making them insufficient for distinguishing samples within the homogeneous modality in VIMI-ReID.

To address this, we introduce the \textbf{D}ivergence \textbf{T}ransformer \textbf{M}odule (DTM) in the 3rd training stage, which learns modality-specific features that capture modality-style information, i.e., the low-level visual characteristics that exist within each modality due to their distinct imaging mechanisms, including color, grayscale spectrum, contrast, and saturation~\cite{Ren2024Implicit}. 
This type of information is inherently associated with the imaging modality and is not directly suitable for heterogeneous-modality matching, but can provide complementary cues to enhance discrimination within the homogeneous modality.
The parameters of STM are frozen during this stage, so that the shared features serve as a stable reference signal for the modality-specific branch to learn complementary information.

After obtaining stable modality-shared features at the end of 2nd stage of training, the images are further processed by the Visible ViT $ViT_{vis}$ and the Infrared ViT $ViT_{ir}$, respectively. These specialized transformers extract modality-specific features tailored to visible and infrared images. The extraction process is formally defined as follows:
\begin{equation}
    F_{sp}^{vis}=ViT_{vis}(x_{vis}),F_{sp}^{ir}=ViT_{ir}(x_{ir}).
    \label{eq6}
\end{equation}

To further obtain modality-style information that is ignored by the shared branch, we design the Divergence Loss $L_D$---between modality-shared and modality-specific features, which reduces the similarity between modality-shared and modality-specific features, thereby steering the modality-specific branch toward learning style information.
The implementation is as follows:
\begin{equation}
\begin{aligned}
    L_D^{vis} =  \frac{\|F_{sh}^{vis} {F_{sp}^{vis}}^T\|_F^2}{B}   ,\
    L_D^{ir} =  \frac{\|F_{sh}^{ir} {F_{sp}^{ir}}^T\|_F^2}{B} .
\end{aligned} 
\label{eq7}
\end{equation}
Here, $L_D^{vis}$ and $L_D^{ir}$ represent difference constraints for visible-specific features and infrared-specific features, respectively. $F_{sh}^{vis}$ and ${F}_{sh}^{ir}$ are the modality-shared features. $\|\mathbf{M}\|_F$ represents the Frobenius norm of matrix $\mathbf{M}$, which measures the magnitude of the matrix elements. $B$ denotes the batch size, indicating the number of samples processed in each training iteration.

The loss function for the modality-specific features in DTM is formulated as follows:
\begin{equation}
    L_{DTM}=L_D^{vis}+L_D^{ir}.
    \label{eq8}
\end{equation}
\subsection{Modality Adaptive Matching Module}
\label{3.4}

VIMI-ReID encompasses both homogeneous-modality and heterogeneous-modality matching tasks. Relying solely on modality-shared features can lead to natural similarity between homogeneous-modality samples interfering with heterogeneous-modality matching. Conversely, using only modality-specific features may result in identity information loss during heterogeneous-modality matching. To address these issues, we propose the \textbf{M}odality \textbf{A}daptive \textbf{M}atching Module (MAM), which adaptively fuses modality-specific and modality-shared features according to the query-gallery modality relationship. 
For homogeneous-modality matching, modality-specific features are used to compensate for the modality-style information ignored in modality-shared features, thereby enhancing homogeneous-modality discrimination.
For heterogeneous-modality matching, MAM relies only on modality-shared features, since modality-style information is modality-dependent and may introduce noise into heterogeneous-modality matching.

Suppose the given query image $Q^{vis}$ belongs to the visible image, while the gallery samples contain both visible samples $G^{vis}$ and infrared samples $G^{ir}$. The modality-shared features and modality-specific features of the query and gallery samples can be extracted through $ViT_{sh}$ and $ViT_{vis}$/$ViT_{ir}$. The specific extraction method is as follows:
\begin{equation}
\begin{alignedat}{2}
    q_{sh}^{vis},  g_{sh}^{vis},  g_{sh}^{ir} &= ViT_{sh}(Q^{vis},G^{vis},G^{ir}) \\
    q_{sp}^{vis},  g_{sp}^{vis}  &= ViT_{vis}(Q^{vis},G^{vis})
\end{alignedat}
.
\label{eq9}
\end{equation}
Here, $q$ and $g$ indicate whether the feature belongs to a query sample or a gallery sample, $vis$ and $ir$ denote whether the feature belongs to the visible or infrared modality, $sh$ represents the modality-shared feature, and $sp$ represents the modality-specific feature. Then, perform modality adaptive fusion of features by determining the matching mode based on the modality labels of the samples. Taking the query as an example, if both the query sample and the gallery sample belong to the visible modality, the feature fusion is performed as follows:
\begin{equation}
    q_{fusion} = q_{sh}^{vis} + \alpha \cdot q_{sp}^{vis},
    \label{eq10}
\end{equation}
where $\alpha$ adaptively controls the contribution of modality-specific features according to the shared-feature similarity. A higher similarity yields a larger $\alpha$, introducing more modality-style information to distinguish highly similar homogeneous-modality samples. It is computed as follows:
\begin{equation}
\alpha =  \frac{1}{1 + \exp(-\operatorname{Sim}(q_{\text{sh}}, g_{\text{sh}}))},
\end{equation}
where $\mathrm{Sim}(\cdot, \cdot)$ denotes the similarity between features.
The similarity calculation between query sample and each gallery sample can be expressed as:
\begin{equation}
    Sim(q,g) = \left\{ 
    \begin{array}{ll}
        q_{fusion} \cdot g_{fusion}, & m_q = m_g \\
        q_{sh} \cdot g_{sh}, & m_q \neq m_g
    \end{array}
    \right.
    .
    \label{eq11}
\end{equation}
Here, $m_q$ and $m_g$ denote the modality labels of the query sample and gallery sample, respectively.
The modality label can generally be inferred from the image acquisition device.

\begin{figure}[t]
    \centering
    \includegraphics[width=1\linewidth]{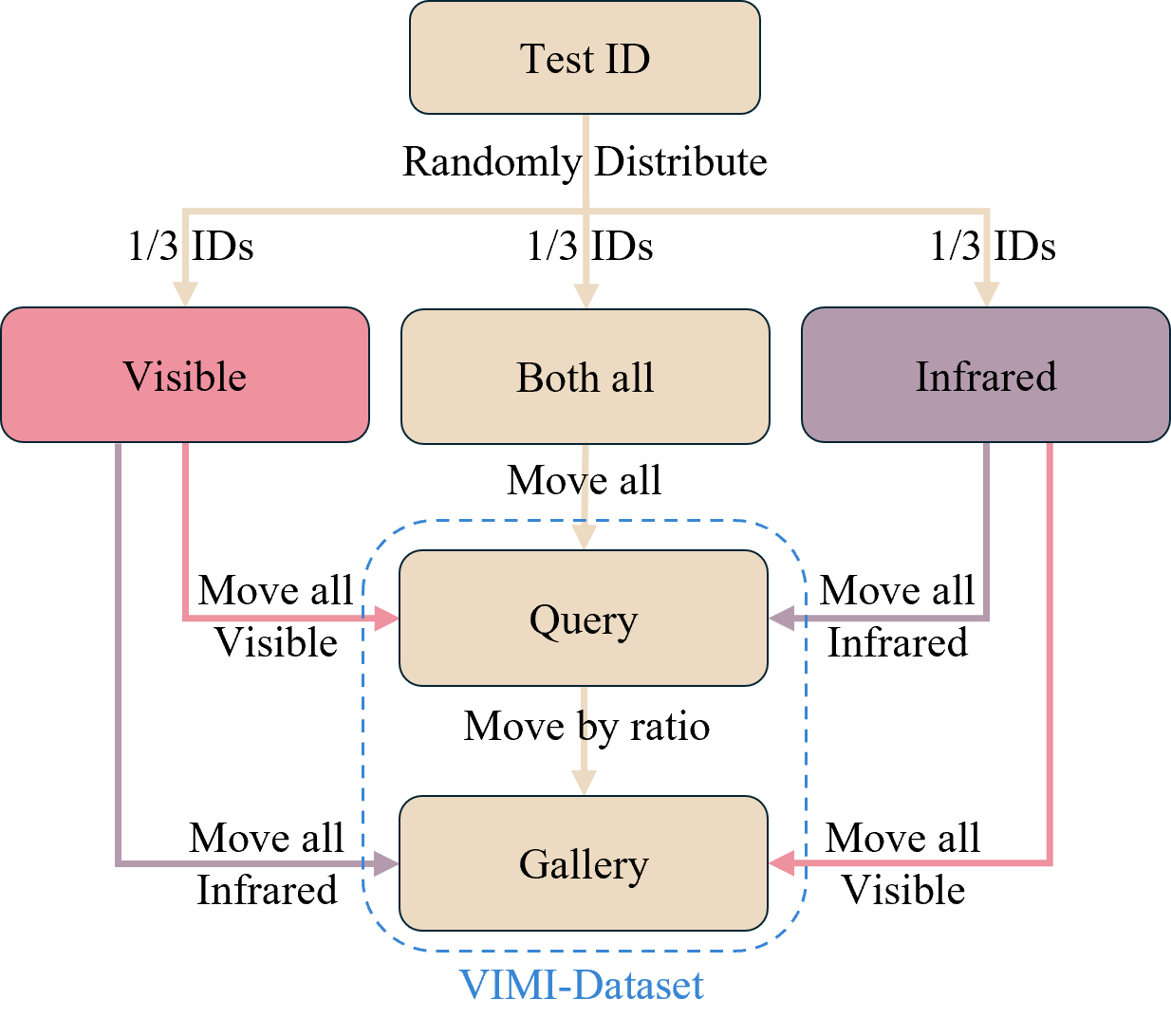}\vspace*{-6mm}
    \caption{Arrows of different colors indicate the movement directions of images from different modalities. The ratio annotation in the figure represents the process of randomly selecting half of the images from identities that contain both visible and infrared images and moving them to the gallery set. For identities that contain only visible or only infrared images, images are moved according to a manually adjustable proportion.}\vspace*{-3mm}
    \label{Dataset}
\end{figure}

\section{Experiments}
\subsection{VIMI-ReID Dataset}
\label{4.1}
The two most widely used datasets in the VI-ReID are SYSU-MM01~\cite{Wu2017RGB} and RegDB~\cite{Nguyen2017Person}. The SYSU-MM01 dataset contains 491 unique identities, with a total of 30,071 visible images and 15,792 infrared images, collected from six cameras under varying illumination conditions. Similarly, the RegDB dataset includes 412 identities, each captured in both modalities, with 4,120 visible and 4,120 infrared images. However, both datasets follow a closed-world evaluation protocol, where the query and gallery images come from heterogeneous modalities.

To meet the requirements of the VIMI-ReID task, we reconstruct the VI-ReID datasets, SYSU-MM01 and RegDB, as SYSU-VIMI and RegDB-VIMI. All test identities are evenly categorized into three groups---those containing visible-only, infrared-only, or both. Images from identities meeting these conditions are assigned to the query set, while the remaining images are placed in the gallery set. Additionally, for identities in the query set that contain both visible and infrared images, half of their images are randomly moved to the gallery set. For identities that contain only visible or only infrared images, a predefined proportion of their images is also transferred to the gallery set. 
A larger proportion means that the numbers of the two modality images in the gallery become closer, resulting in a more severe degree of homogeneous-modality interference.

The detailed dataset reconstruction process is shown in Figure~\ref{Dataset}. The identities in the test set are divided into three categories: visible-only, infrared-only, and both.
For visible-only identities, all infrared images are moved to the gallery set, while all visible images are moved to the query set, with a portion of them (according to a predefined ratio) also moved to the gallery set.
For infrared-only identities, the process is reversed: all visible images are moved to the gallery set, and all infrared images are moved to the query set, with a portion of them moved to the gallery set as well.
For identities in the both category, all images are first moved to the query set, and then half of the visible and half of the infrared images are randomly selected and added to the gallery set.

\begin{table}[t]
\caption{Comparison of the number of identities before and after dataset reconstruction. The two testing settings of RegDB are denoted in parentheses.}\vspace*{-2mm}
\label{Reconstruction}
\centering
\setlength{\tabcolsep}{4pt} 
\begin{tabular}{@{}lcccccc@{}}
\toprule
\multirow{2}{*}{Testing Dataset} 
& \multicolumn{3}{c}{SYSU-MM01} 
& \multicolumn{3}{c}{RegDB} \\ 
\cmidrule(lr){2-4}\cmidrule(lr){5-7}
& Vis & Both & IR & Vis & Both & IR \\
\midrule
Query ID   & 96  & - & - & (206) & - & 206 \\
Gallery ID & -  & - & 96 & 206   & - & (206) \\
\midrule
\multicolumn{1}{l}{}
& \multicolumn{3}{c}{SYSU-VIMI} 
& \multicolumn{3}{c}{RegDB-VIMI} \\ 
\cmidrule(lr){2-4}\cmidrule(lr){5-7}
Query ID   & 32 & 32 & 32 & 68 & 69 & 69 \\
Gallery ID & - & 96 & - & - & 206 & - \\
\botrule
\end{tabular}\vspace*{-4mm}
\end{table}

\begin{figure}[t]
    \centering
    \includegraphics[width=1\linewidth]{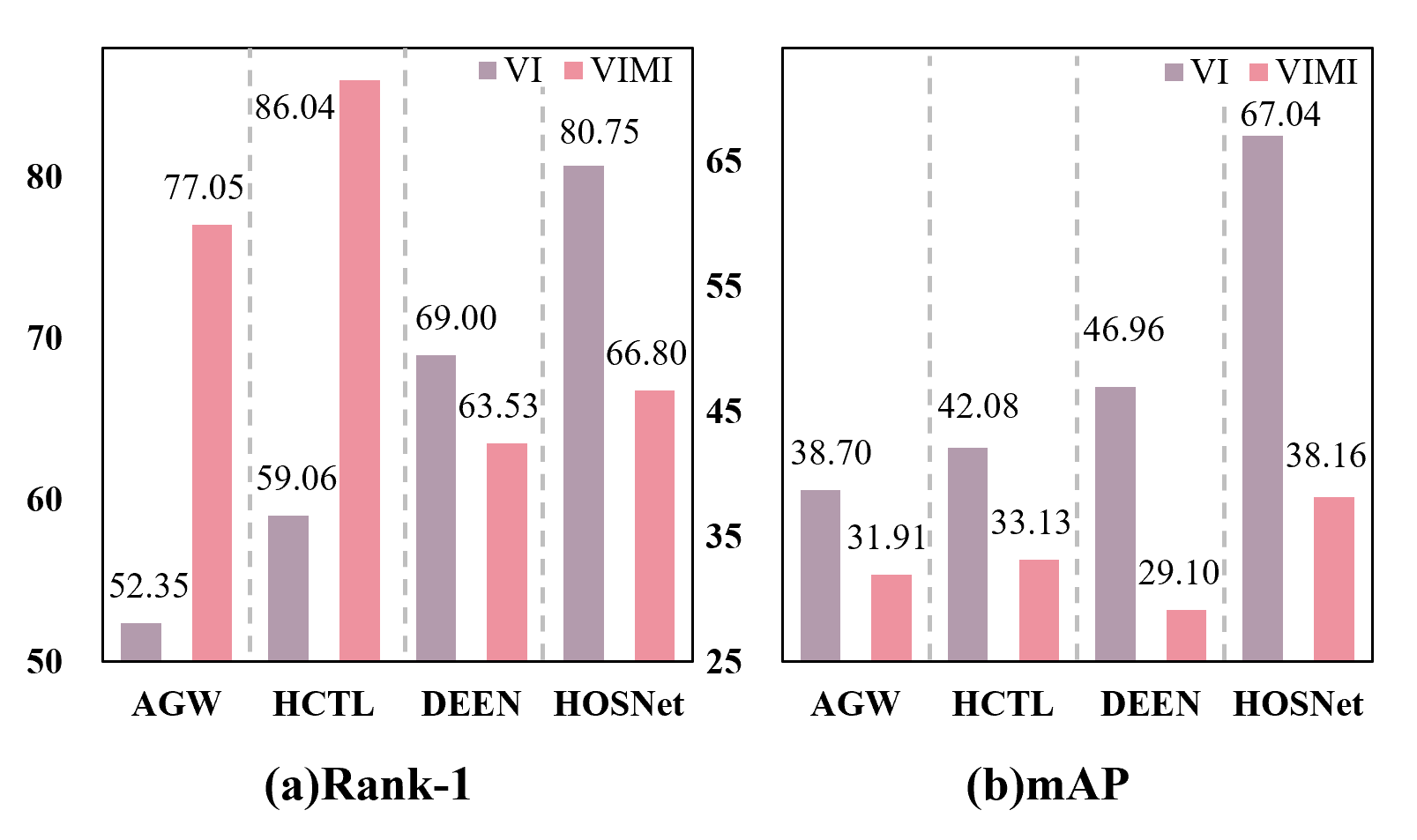}\vspace*{-6mm}
    \caption{Comparison of Mainstream VI-ReID Methods in the VI-ReID and VIMI-ReID Tasks.}\vspace*{-3mm}
    \label{drop}
\end{figure}

The changes in the reconstructed datasets can be observed in Table~\ref{Reconstruction}. As shown in the table, SYSU-MM01 and RegDB provide only one type of heterogeneous-modality retrieval setting (either Vis-to-IR or IR-to-Vis). In contrast, the reconstructed datasets enable each query sample to retrieve images from both modalities simultaneously.
The gallery contains samples from both modalities to better simulate an open-world retrieval scenario. This new partitioning scheme produces four retrieval settings, namely VIMI-Homo, VIMI-Hetero, VIMC, and VIMI.
Specifically, VIMI-Homo refers to retrieval between images of the homogeneous modality (IR-to-IR and Vis-to-Vis), which resembles single-modality retrieval but additionally includes infrared-based retrieval compared with traditional V-ReID. VIMI-Hetero denotes retrieval between heterogeneous modalities (IR-to-Vis and Vis-to-IR). VIMI-Hetero differs slightly from conventional VI-ReID, as VIMI-Hetero covers both heterogeneous-modality directions while still reflecting the performance gap under typical VI-ReID conditions. VIMC represents the ideal scenario in an open-world setting, as it includes both homogeneous-modality and heterogeneous-modality retrieval. In this setting, each query identity has both visible and infrared images in the gallery. In contrast, VIMI simulates a more realistic open-world scenario in which the gallery for some identities becomes modality-incomplete. 
In the VIMI setting, the gallery composition consists of three types of identities: 25\% retain only visible images, 25\% retain only infrared images, and the remaining 50\% retain images from both modalities.

\begin{table*}[th]
\caption{Comparison with SOTA methods on the RegDB-VIMI dataset under mixing ratio 0.5. The best and second-best results are highlighted in \textbf{bold} and \underline{underline}, respectively.}
\label{RegDB_05}
\small
\centering
\begin{tabularx}{\textwidth}{lcCCCCCCCC}
\toprule
\multirow{2}{*}{Method} & \multirow{2}{*}{Venue} 
    & \multicolumn{2}{c}{VIMI} 
    & \multicolumn{2}{c}{VIMI-Homo} 
    & \multicolumn{2}{c}{VIMI-Hetero} 
    & \multicolumn{2}{c}{VIMC} \\
\cmidrule(lr){3-4}
\cmidrule(lr){5-6}
\cmidrule(lr){7-8}
\cmidrule(lr){9-10}
&  & Rank1 & mAP & Rank1 & mAP & Rank1 & mAP & Rank1 & mAP\\
\midrule
HCTL~\cite{Liu2021Parameter}  & TMM'20 
    & 81.87 & 62.28 & 91.17 & 83.13 & 58.67 & 50.35 & 93.48 & 62.04\\
    
AGW~\cite{Ye2022Deep}   & TPAMI'21 
    & 76.14 & 60.99 & 87.88 & 82.10 & 52.52 & 50.69 & 86.19 & 60.44\\

PMT~\cite{Lu2023Learning}   & AAAI'23  
    & \underline{96.65} & \underline{92.08} 
    & \underline{99.83} & 98.06 
    & 85.41 & 80.64 
    & \underline{99.84} & \underline{84.46}\\

DEEN~\cite{Zhang2023Diverse}  & CVPR'23  
    & 92.91 & 84.55 
    & 99.55 & 98.03 
    & \underline{85.63} & \underline{81.62} 
    & 99.53 & 83.36 \\

HOSNet~\cite{Qiu2024High} & AAAI'24 
    & 85.91 & 72.05 
    & 95.55 & 89.45 
    & 69.79 & 65.95 
    & 94.89 & 69.84\\

AMML~\cite{zhang2025adaptive} & IJCV'25 
    & 87.49 & 73.60 
    & 99.82 & \underline{98.41} 
    & 78.49 & 73.93 
    & \underline{99.84} & 71.49\\

IRL~\cite{Wang2025Low} &
ACM MM'25 &
92.79 &
81.61 &
99.69 &
97.22 &
82.32 &
74.73 &
99.73 &
80.06 \\

\midrule
Ours  & -- 
    & \textbf{97.81} & \textbf{94.03} 
    & \textbf{99.97} & \textbf{99.37} 
    & \textbf{94.21} & \textbf{92.00} 
    & \textbf{99.97} & \textbf{93.52} \\
\botrule
\end{tabularx}\vspace*{-4mm}
\end{table*}

\begin{table*}[th]
\caption{Comparison with SOTA methods on the RegDB-VIMI dataset under mixing ratio 0.1.
}
\label{RegDB_01}
\small
\centering
\begin{tabularx}{\textwidth}{lcCCCCCCCC}
\toprule
\multirow{2}{*}{Method}  & \multirow{2}{*}{Venue}  
    & \multicolumn{2}{c}{VIMI} 
    & \multicolumn{2}{c}{VIMI-Homo} 
    & \multicolumn{2}{c}{VIMI-Hetero} 
    & \multicolumn{2}{c}{VIMC} \\
\cmidrule(lr){3-4} 
\cmidrule(lr){5-6} 
\cmidrule(lr){7-8} 
\cmidrule(lr){9-10}
&  & Rank1 & mAP & Rank1 & mAP & Rank1 & mAP & Rank1 & mAP \\
\midrule
HCTL~\cite{Liu2021Parameter}  & TMM'20 
    & 75.73 & 59.51 
    & 81.38 & 82.28 
    & 59.39 & 51.89 
    & 86.63 & 55.71 \\

AGW~\cite{Ye2022Deep}   & TPAMI'21 
    & 70.53 & 57.55 
    & 79.36 & 81.10 
    & 54.25 & 52.77 
    & 80.32 & 54.71 \\

PMT~\cite{Lu2023Learning}   & AAAI'23  
    & \underline{96.10} & \underline{91.12} 
    & 96.94 & 97.34 
    & \underline{86.55} & \underline{82.70} 
    & 98.28 & \underline{80.28} \\

DEEN~\cite{Zhang2023Diverse}  & CVPR'23  
    & 91.41 & 82.19 
    & 97.09 & 97.26 
    & 83.91 & 81.21 
    & 97.49 & 77.35 \\

HOSNet~\cite{Qiu2024High} & AAAI'24 
    & 82.24 & 70.00 
    & 87.66 & 88.32 
    & 70.34 & 67.58 
    & 90.66 & 65.55 \\

AMML~\cite{zhang2025adaptive} & IJCV'25 
    & 86.83 & 70.53 
    & \underline{97.92} & \underline{97.91} 
    & 78.53 & 74.35 
    & \underline{98.74} & 67.64 \\

IRL~\cite{Wang2025Low} &
ACM MM'25 &
91.98 &
79.41 &
96.58 &
96.53 &
83.04 &
76.48 &
98.29 &
74.99 \\

\midrule
Ours  & -- 
    & \textbf{97.38} & \textbf{93.36} 
    & \textbf{98.97} & \textbf{99.07} 
    & \textbf{94.50} & \textbf{92.71} 
    & \textbf{99.44} & \textbf{91.31} \\
\botrule
\end{tabularx}\vspace*{-4mm}
\end{table*}

The reconstruction introduces a substantial number of homogeneous-modality samples into the retrieval process, significantly increasing task complexity and posing new challenges for existing VI-ReID methods, as illustrated in Figure~\ref{drop}. Empirical results demonstrate that most VI-ReID models experience a pronounced performance decline on the VIMI-ReID datasets compared to their performance on the original ones. 
This degradation indicates that existing methods are insufficient to effectively address the VIMI-ReID task.

\subsection{Experiment Setting}
\textbf{Evaluation protocol.}
\textbf{C}umulative \textbf{M}atching \textbf{C}haracteristics (CMC) and \textbf{m}ean \textbf{A}verage \textbf{P}recision (mAP) are used as the evaluation metrics in our experiments.
We select representative VI-ReID methods published in recent years for comparison, including AGW~\cite{Ye2022Deep}, HCTL~\cite{Liu2021Parameter}, DEEN~\cite{Zhang2023Diverse}, PMT~\cite{Lu2023Learning}, HOS-Net~\cite{Qiu2024High}, AMML~\cite{zhang2025adaptive}, DiVE~\cite{dai2025diffusion}, and IRL~\cite{Wang2025Low}.
All compared methods are trained following the strategies reported in their original papers to ensure consistency with the original results and maintain a fair comparison.
For testing on SYSU-VIMI, the original testing protocol was modified to align with the VIMI-ReID paradigm by replacing the random selection of gallery images with the entire gallery set for evaluation. For RegDB-VIMI, the full gallery set was also used for testing, with results obtained by averaging performance over 10 test runs.

\begin{table*}[th]
\caption{Comparison with SOTA methods on the SYSU-VIMI dataset under mixing ratio 0.5. }
\small
\label{SYSU_05}
\centering
\begin{tabularx}{\textwidth}{lcCCCCCCCC}
\toprule
\multirow{2}{*}{Method} & \multirow{2}{*}{Venue} 
    & \multicolumn{2}{c}{VIMI} 
    & \multicolumn{2}{c}{VIMI-Homo} 
    & \multicolumn{2}{c}{VIMI-Hetero} 
    & \multicolumn{2}{c}{VIMC}\\
\cmidrule(lr){3-4}
\cmidrule(lr){5-6}
\cmidrule(lr){7-8}
\cmidrule(lr){9-10}
& & Rank1 & mAP & Rank1 & mAP & Rank1 & mAP & Rank1 & mAP\\
\midrule
HCTL~\cite{Liu2021Parameter} & TMM'20
    & 76.72 & 44.34 & 95.71 & 67.86 & 39.47 & 27.55 & 95.37 & 44.30\\

AGW~\cite{Ye2022Deep} & TPAMI'21 
    & 76.25 & 46.13 & 88.61 & 62.39 & 34.91 & 26.52 & 86.68 & 42.56\\

PMT~\cite{Lu2023Learning} & AAAI'23 
    & \underline{85.10} & \underline{66.71} & \textbf{99.27} & \textbf{87.39} & \textbf{68.72} & \underline{53.54} & \textbf{99.31} & \underline{67.24}\\
DEEN~\cite{Zhang2023Diverse} & CVPR'23 
    & 65.61 & 32.90 & 77.21 & 42.22 & 39.66 & 26.43 & 77.02 & 33.08\\
HOSNet~\cite{Qiu2024High} & AAAI'24 
    & 62.84 & 40.43 & 71.80 & 48.41 & 47.99 & 37.80 & 73.63 & 42.38\\
AMML~\cite{zhang2025adaptive} & IJCV'25 
    & 72.17 & 36.09 & 86.58 & 50.50 & 40.48 & 26.75 & 86.36 & 35.56\\

DiVE~\cite{dai2025diffusion} &
AAAI'25 &
70.91 &
45.79 &
81.48 &
56.62 &
54.33 &
40.38 &
83.06 &
46.92 \\

IRL~\cite{Wang2025Low} &
ACM MM'25 &
81.14 &
52.96 &
95.84 &
70.23 &
59.28 &
43.06 &
95.58 &
54.84 \\

\midrule
Ours & -- 
    & \textbf{91.32} & \textbf{71.50} 
    & \underline{99.01} & \underline{86.71} 
    & \underline{66.87} & \textbf{54.30} 
    & \underline{98.91} & \textbf{69.50}\\
\botrule
\end{tabularx}
\vspace*{-4mm}
\end{table*}

\begin{table*}[th]
\caption{Comparison with SOTA methods on the SYSU-VIMI dataset under mixing ratio 0.1. 
}
\label{SYSU_01}
\small
\centering
\begin{tabularx}{\textwidth}{lcCCCCCCCC}
\toprule
\multirow{2}{*}{Method} & \multirow{2}{*}{Venue} 
    & \multicolumn{2}{c}{VIMI} 
    & \multicolumn{2}{c}{VIMI-Homo} 
    & \multicolumn{2}{c}{VIMI-Hetero} 
    & \multicolumn{2}{c}{VIMC} \\
\cmidrule(lr){3-4}
\cmidrule(lr){5-6}
\cmidrule(lr){7-8}
\cmidrule(lr){9-10}
&  & Rank1 & mAP & Rank1 & mAP & Rank1 & mAP & Rank1 & mAP \\
\midrule
HCTL~\cite{Liu2021Parameter}  & TMM'20 
    & 71.66 & 38.05 & 85.11 & 58.55 & 43.34 & 31.81 & 86.04 & 33.13 \\
AGW~\cite{Ye2022Deep}   & TPAMI'21 
    & 69.19 & 36.77 & 76.16 & 52.30 & 40.06 & 29.89 & 77.05 & 31.91 \\

PMT~\cite{Lu2023Learning}   & AAAI'23  
    & \textbf{91.25} & \underline{64.95} 
    & \underline{96.14} & \textbf{80.93} 
    & \textbf{71.88} & \underline{58.85} 
    & \underline{96.22} & \underline{55.78} \\
DEEN~\cite{Zhang2023Diverse}  & CVPR'23  
    & 53.92 & 28.89 & 59.50 & 35.48 & 41.49 & 30.12 & 63.53 & 29.10 \\
HOSNet~\cite{Qiu2024High} & AAAI'24 
    & 57.35 & 37.55 & 60.92 & 43.13 & 48.77 & 40.85 & 66.80 & 38.16 \\
AMML~\cite{zhang2025adaptive} & IJCV'25 
    & 61.13 & 30.66 & 71.46 & 44.10 & 41.53 & 29.48 & 72.96 & 28.96 \\

DiVE~\cite{dai2025diffusion} &
AAAI'25 &
66.13 &
41.38 &
71.29 &
50.70 &
55.43 &
43.70 &
77.14 &
41.91 \\

IRL~\cite{Wang2025Low} &
ACM MM'25 &
77.27 &
48.03 &
86.97 &
63.69 &
62.60 &
47.08 &
89.29 &
45.96 \\
    
\midrule
Ours  & -- 
    & \underline{89.62} & \textbf{65.94} 
    & \textbf{96.16} & \underline{80.91} 
    & \underline{71.53} & \textbf{60.30} 
    & \textbf{96.37} & \textbf{60.34} \\
\botrule
\end{tabularx}
\vspace*{-4mm}
\end{table*}

\textbf{Implementation details.} All compared methods and our proposed method are implemented and evaluated on a single NVIDIA RTX 3090 GPU.
We adopt ViT-B/16~\cite{Dosovitskiy2021Image}, pretrained on ImageNet~\cite{Deng2009ImageNet}, as the backbone network. The overlap stride is set to 12 to balance computational efficiency and recognition performance.
All person images are resized to $256 \times 128$ pixels. Horizontal flipping and random erasing are used for data augmentation. For infrared images, color jitter and Gaussian blur are additionally applied.
The batch size is set to 64, and each mini-batch contains 8 identities. For each identity, we sample 4 visible images and 4 infrared images.
We use the AdamW optimizer with a cosine annealing learning rate scheduler. The initial learning rate is set to $3 \times 10^{-4}$, and the weight decay is set to $1 \times 10^{-4}$.
The model is trained for 36 epochs on SYSU-MM01 and 64 epochs on RegDB.
For both datasets, the first-stage training epoch is set to 6. The second-stage training epoch is set to 24 on SYSU-MM01 and 36 on RegDB. During testing, the 768-dimensional features after the BN layer are used for retrieval.


\subsection{Comparison with State-of-the-art Methods}
In this subsection, we compare MAMT with state-of-the-art methods on the SYSU-VIMI and RegDB-VIMI benchmarks.
DiVE is reported only on SYSU-VIMI because its original data-generation procedure is not available for RegDB.

\textbf{RegDB-VIMI:} As shown in Tables~\ref{RegDB_05} and~\ref{RegDB_01}, our method consistently achieves the best performance across all evaluation settings. On the RegDB-VIMI (0.5) dataset, our method surpasses the second-best method by \textbf{0.96\%} in mAP under the VIMI-Homo setting. Under the VIMI-Hetero setting, it achieves substantial improvements of \textbf{8.58\%} in Rank-1 and \textbf{10.38\%} in mAP. Under the VIMC setting, our method further exceeds the second-best method by \textbf{9.06\%} in mAP. In the modality-incomplete VIMI setting, our method also improves Rank-1 and mAP by \textbf{1.16\%} and \textbf{1.95\%}, respectively. On the RegDB-VIMI (0.1) dataset, the performance improvements are consistent with the trends observed on RegDB-VIMI (0.5). Under the VIMI setting, our method achieves gains of \textbf{1.28\%} in Rank-1 and \textbf{2.24\%} in mAP. Under the VIMC setting, it improves Rank-1 by \textbf{0.70\%} and mAP by \textbf{11.03\%}.
These results demonstrate that our method achieves significant improvements in both homogeneous-modality and heterogeneous-modality retrieval, with particularly notable advantages in overall retrieval quality. Moreover, even when faced with mixed-modality and modality-incomplete scenarios, our method maintains strong robustness.

\textbf{SYSU-VIMI:} 
As shown in Tables~\ref{SYSU_05} and~\ref{SYSU_01}, our method outperforms existing methods under most evaluation settings on the SYSU-VIMI dataset. Notably, our method demonstrates a substantial advantage in mAP, indicating its stronger overall retrieval capability. 
In SYSU-VIMI (0.5), our method achieves comparable performance to the best existing methods under both VIMI-Homo and VIMI-Hetero settings, since MAM is mainly designed to handle mixed-modality and modality-incomplete scenarios rather than purely homogeneous or purely heterogeneous settings.
In the VIMC setting, our method shows slightly lower Rank-1 accuracy compared with the best-performing method. This is because MAM may overcorrect some query-gallery pairs, leading to a few improperly adjusted homogeneous-modality matches. Despite this, our method still improves mAP by \textbf{2.26\%}, suggesting that the overall retrieval quality is enhanced even if a small portion of matches are adversely affected.
In the modality-incomplete VIMI setting, our method surpasses the second-best method by \textbf{6.22\%} in Rank-1 and \textbf{4.79\%} in mAP. A similar trend is observed in SYSU-VIMI (0.1), where our method also achieves consistent improvements under both VIMI and VIMC settings, improving mAP by \textbf{0.99\%} and \textbf{4.56\%}, respectively.
These results demonstrate that our method achieves a more desirable balance between top-rank accuracy and overall ranking performance, maintaining Rank-1 performance while delivering significantly higher mAP.

\subsection{Ablation Study}
To assess the contribution of each component and verify the necessity of each module, we perform systematic ablation studies on the SYSU-VIMI (0.1) by progressively incorporating key components to evaluate their individual and combined impact on overall performance. Given that VIMI is generated through random sampling and therefore contains inherent uncertainty, all ablation studies are conducted under the VIMC setting.

\textbf{Effect of Grayscale Conversion on VIMI-ReID.} To further explore the impact of grayscale conversion on the VIMI-ReID task, we compare our method with recent pseudo-modality approaches, MMN~\cite{Zhang2021Towards} and SGIEL~\cite{Feng2023Shape}, as shown in Table~\ref{tab2}. When grayscale conversion is not applied (RGB setting), our method already achieves substantially higher performance than pseudo-modality methods, surpassing SGIEL by \textbf{5.23\%} in Rank-1 and \textbf{9.61\%} in mAP. When only grayscale images are used (Gray setting), Rank-1 remains comparable at 95.53\% but mAP drops drastically to 34.69\%, a decrease of \textbf{21.10\%} compared to the RGB setting. This suggests that using only grayscale images reduces the modality discrepancy but weakens the discriminability of the learned shared features. The Gray-RGB setting, which introduces grayscale conversion only in Stage 1 before transitioning to RGB setting, achieves the best performance on both metrics with Rank-1 of \textbf{96.37\%} and mAP of \textbf{60.34\%}. 
This demonstrates that grayscale conversion is beneficial for stabilizing early modality-shared feature learning, while subsequent RGB training is necessary to preserve the discriminability of the final shared representations.

\begin{table}[t]
\caption{Effect of grayscale conversion on VIMI-ReID.}
\label{tab2}
\centering
\setlength{\tabcolsep}{4mm}
\begin{tabular}{lccc}
\toprule
Mode & Method & \multicolumn{1}{c}{Rank-1} & \multicolumn{1}{c}{mAP} \\
\midrule
Pseudo  
 & MMN~\cite{Zhang2021Towards}   & 72.54 & 29.02 \\
 -modality & SGIEL~\cite{Feng2023Shape} & 89.86 & 46.18 \\
\midrule
\multirow{3}{*}{Grayscale}       
 & RGB       & 95.09 & \underline{55.79} \\
 & Gray      & \underline{95.53}  & 34.69  \\
 & Gray-RGB  & \textbf{96.37}  & \textbf{60.34}  \\
\bottomrule
\end{tabular}\vspace*{-4mm}
\end{table}

\begin{table}[t]
\caption{Effect of MAM and $L_D$ on MAMT under different backbones.}
\label{tab3}
\centering
\setlength{\tabcolsep}{2.8mm}
\begin{tabular}{lcccccc}
\toprule
Backbone & Vis & IR & $L_D$ & \multicolumn{1}{c}{Rank-1} & \multicolumn{1}{c}{mAP} \\
\midrule
\multirow{7}{*}{ResNet}  
 &  &  &  &  89.20 & 37.35  \\
  & \checkmark &  &  & 91.69  & 40.13 \\
  &  & \checkmark &  & 89.16 & 39.41  \\
 & \checkmark & \checkmark &  & 90.10 & 41.19  \\
 & \checkmark &  & \checkmark & 92.11  & 38.35 \\
 &  & \checkmark & \checkmark & 90.55  & 39.47  \\
 & \checkmark & \checkmark & \checkmark &\textbf{92.76}  & \textbf{41.59} \\
\midrule
\multirow{7}{*}{ViT}  
 &  &  &  & 94.29  & 56.32  \\
  & \checkmark &  &  & 95.33  & 57.73 \\
  &  & \checkmark &  & 94.96 & 56.99  \\
 & \checkmark & \checkmark &  & \textbf{96.37} & 59.39  \\
 & \checkmark &  & \checkmark & 95.83  & 58.72 \\
 &  & \checkmark & \checkmark & 95.22  & 57.93  \\
 & \checkmark & \checkmark & \checkmark & \textbf{96.37}  & \textbf{60.34} \\
\bottomrule
\end{tabular}\vspace*{-4mm}
\end{table}

\begin{table*}[t]
\caption{Comparison of MAMT with Multi-model Ensemble methods and All-in-One methods. \textbf{Bold} denotes the best performance under each corresponding evaluation setting.}
\label{modal}
\small
\centering
\begin{tabularx}{\textwidth}{lcccCCCC}
\toprule
\multirow{2}{*}{Method} &\multirow{2}{*}{Mode} &\multirow{2}{*}{Train} & \multirow{2}{*}{Test} & \multicolumn{2}{c}{Performance} & \multicolumn{2}{c}{Cost} \\
\cmidrule(lr){5-6} \cmidrule(lr){7-8}
 & & & & Rank-1 & mAP & Memory & Time \\
\midrule
\multirow{5}{*}{AGW~\cite{Ye2022Deep}}
&\multirow{5}{*}{
    \parbox[c]{2.2cm}{\centering Multi-model\\Ensemble}
} & Vis-only & Vis-only & 89.80 & \textbf{85.10} & \multirow{5}{*}{20G} & \multirow{5}{*}{8h} \\
& & IR-only & IR-only  & 77.90 & 73.50 &  &  \\
& & Vis-only & VIMI-Homo & 47.70 & 26.30 &  &  \\
& & Vis\&IR & VIMI-Hetero & 40.06 & 29.89 &  & \\
& &- &VIMC &- &- & & \\
\midrule
\multirow{3}{*}{IRM~\cite{He_2024_CVPR}}
&\multirow{3}{*}{
    \parbox[c]{2.2cm}{\centering All-in-\\One}
} & \multirow{3}{*}{-} & VIMI-Homo  & 82.40 & 44.70 & \multirow{3}{*}{-} & \multirow{3}{*}{-} \\
& & & VIMI-Hetero & 27.10 & 16.90 & & \\
& & & VIMC   & 82.50 & 16.50 & & \\
\midrule
\multirow{5}{*}{Ours}
&\multirow{5}{*}{-} & \multirow{5}{*}{Vis\&IR} & Vis-only    & \textbf{96.67} & 78.94 & \multirow{5}{*}{12G} & \multirow{5}{*}{5h} \\
& &  & IR-only     & \textbf{93.57} & \textbf{79.55} &  &  \\
& &  & VIMI-Homo   & \textbf{96.16} & \textbf{80.91} &  &  \\
& & & VIMI-Hetero & \textbf{71.53} & \textbf{60.30} &  &  \\
& & & VIMC &\textbf{96.37} &\textbf{60.34} & & \\

\botrule
\end{tabularx}
\vspace*{-4mm}
\end{table*}

\textbf{Effect of the Backbone.}
To verify the generalizability of the proposed modules across different architectures, we evaluate them under both ResNet and ViT backbones, as shown in Table~\ref{tab3}. Under ResNet, the full model \textbf{(Row 7)} improves Rank-1 by \textbf{3.56\%} and mAP by \textbf{4.24\%} over the ResNet baseline \textbf{(Row 1)}. Under ViT, the full model \textbf{(Row 14)} improves Rank-1 by \textbf{2.08\%} and mAP by \textbf{4.02\%} over the ViT baseline \textbf{(Row 8)}. The consistent improvements across both backbones confirm the architecture-agnostic effectiveness of the proposed modules, while the overall superior performance under ViT reflects the stronger representational capacity of the Transformer architecture.

\textbf{Effect of MAM.} As shown in \textbf{Rows 8-11} of Table~\ref{tab3}, under the ViT backbone, adding only the visible-specific or infrared-specific branch yields moderate improvements of \textbf{1.04\%}/\textbf{1.41\%} and \textbf{0.67\%}/\textbf{0.67\%} in Rank-1 and mAP, respectively. This indicates that each branch contributes complementary modality-style information, but their individual effects are limited. When both branches are jointly activated, the model achieves more substantial improvements of \textbf{2.08\%} in Rank-1 and \textbf{3.07\%} in mAP, demonstrating that the two modality-specific branches together enhance homogeneous-modality matching more effectively than either branch alone.

\textbf{Effect of the Divergence Loss.} As shown in \textbf{Rows 11 and 14} of Table~\ref{tab3}, introducing $L_D$ on top of both modality-specific branches maintains Rank-1 while further improving mAP by \textbf{0.95\%}. This confirms that $L_D$ effectively enforces disentanglement between shared and specific features, steering the modality-specific branch toward learning genuinely complementary style information rather than duplicating information already captured by the shared branch.

\subsection{Further Analysis}
In this subsection, we compare MAMT with multi-model ensemble methods and all-in-one methods on the VIMI-ReID task, and present the visualization results of MAMT.

\textbf{Comparison with Multi-model Ensemble Methods.}
As shown in \textbf{Rows 1-5} and \textbf{9-13} of Table~\ref{modal}, we compare MAMT with the multi-model ensemble method. We adopt AGW as the baseline and train separate models for different modality retrieval settings. Since separately trained models lack a unified matching strategy for jointly handling homogeneous- and heterogeneous-modality matching, the multi-model ensemble method cannot be directly evaluated under the VIMC setting. The results show that AGW performs well under single-modality retrieval settings (Vis-only and IR-only), but its performance drops substantially under the VIMI-Homo and VIMI-Hetero settings. This indicates that multi-model ensemble methods cannot effectively solve the VIMI-ReID task. Regarding computational resources, the multi-model ensemble strategy requires 20G GPU memory and 8h training time, whereas MAMT requires only 12G GPU memory and 5h training time. This means that the ensemble method consumes about 67\% more memory and requires about 60\% more training time than MAMT. These results indicate that MAMT achieves superior performance with lower computational cost, highlighting the necessity of developing a unified framework for VIMI-ReID.

\begin{figure}[!t]
    \centering
    \includegraphics[width=1\linewidth]{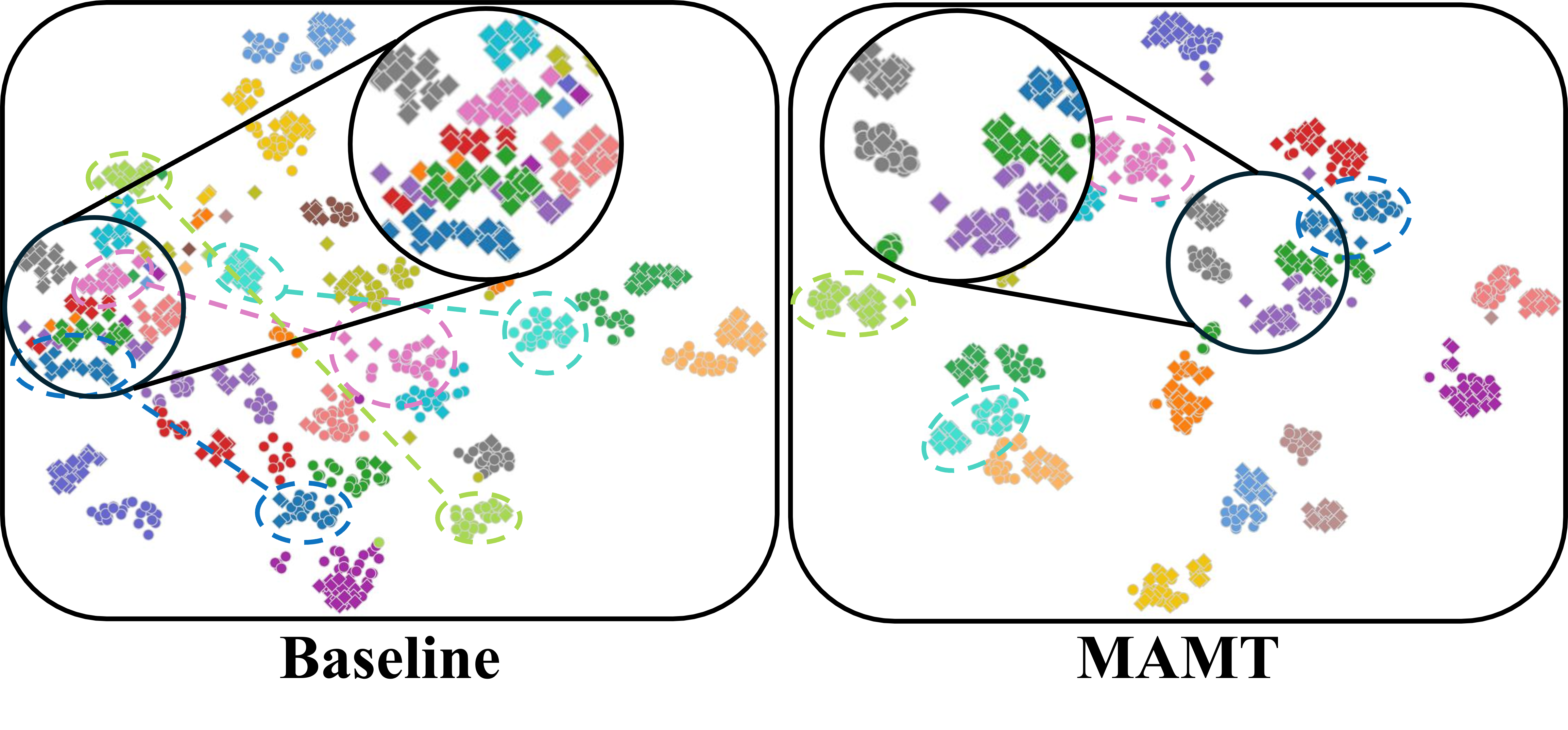}\vspace*{-5mm}
    \caption{The feature distributions of MAMT and the Baseline are visualized. Dashed boxes indicate the aggregation of samples from the same identity, with circles and squares representing visible and infrared images.}\vspace*{-4mm}
    \label{float}
\end{figure}

\textbf{Comparison with All-in-One Method.}
As shown in \textbf{Rows 6-13} of Table~\ref{modal}, we compare MAMT with the all-in-one method IRM~\cite{He_2024_CVPR}. Since IRM is trained with multiple tasks in its original framework, we do not perform additional task-specific training for VIMI-ReID and directly evaluate its performance under different VIMI-ReID settings. The results show that while IRM achieves relatively high Rank-1, its mAP under the VIMC setting is significantly lower than ours due to the interference introduced by homogeneous modalities. This indicates that an all-in-one method is insufficient to address the homogeneous-modality interference specific to VIMI-ReID.

\textbf{Visualization.} 
As shown in Figure~\ref{float}, in the baseline, some identity samples are well classified; however, in the enlarged region, multiple identity clusters become mixed. This phenomenon stems from the inherent homogeneous-modality interference in the VIMI-ReID task, which severely impairs the model’s ability to retrieve heterogeneous-modality samples. In contrast, MAMT shows only minor interference from a few samples, while most identities are properly clustered. This observation demonstrates that MAMT alleviates the issue of homogeneous-modality interference in VIMI-ReID.

\begin{figure}[!t]
    \centering
    \vspace*{-2mm}
    \includegraphics[width=1\linewidth]{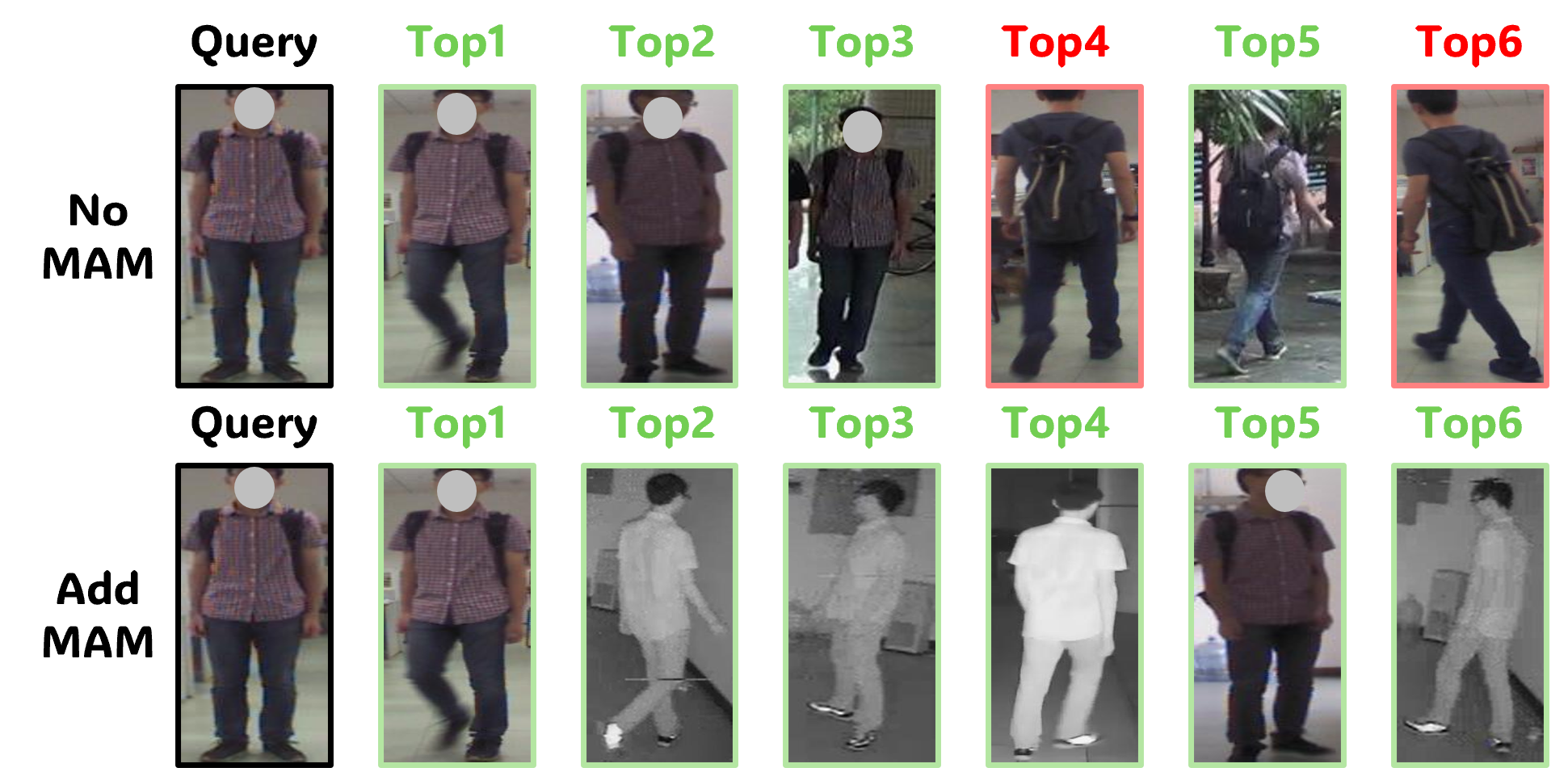}\vspace*{-5mm}
    \caption{Ranking results before and after compensating modality-shared features with modality-specific features.}\vspace*{-5mm}
    \label{figrank}
\end{figure}

As shown in Figure~\ref{figrank}, we visualize the effect of MAM on the ranking results of MAMT. Without MAM, the Top-1 to Top-6 retrieved samples all come from the homogeneous modality as the query. This indicates that modality similarity biases the ranking results, causing some incorrect homogeneous-modality samples to be ranked ahead of correct matches. After introducing MAM, the ranking results are clearly improved: the Top-1 to Top-6 samples include both homogeneous-modality and heterogeneous-modality images with the same identity as the query. This demonstrates that MAM leverages modality similarity to alleviate homogeneous-modality interference.

\begin{figure}[!t]
    \centering
    \includegraphics[width=1\linewidth]{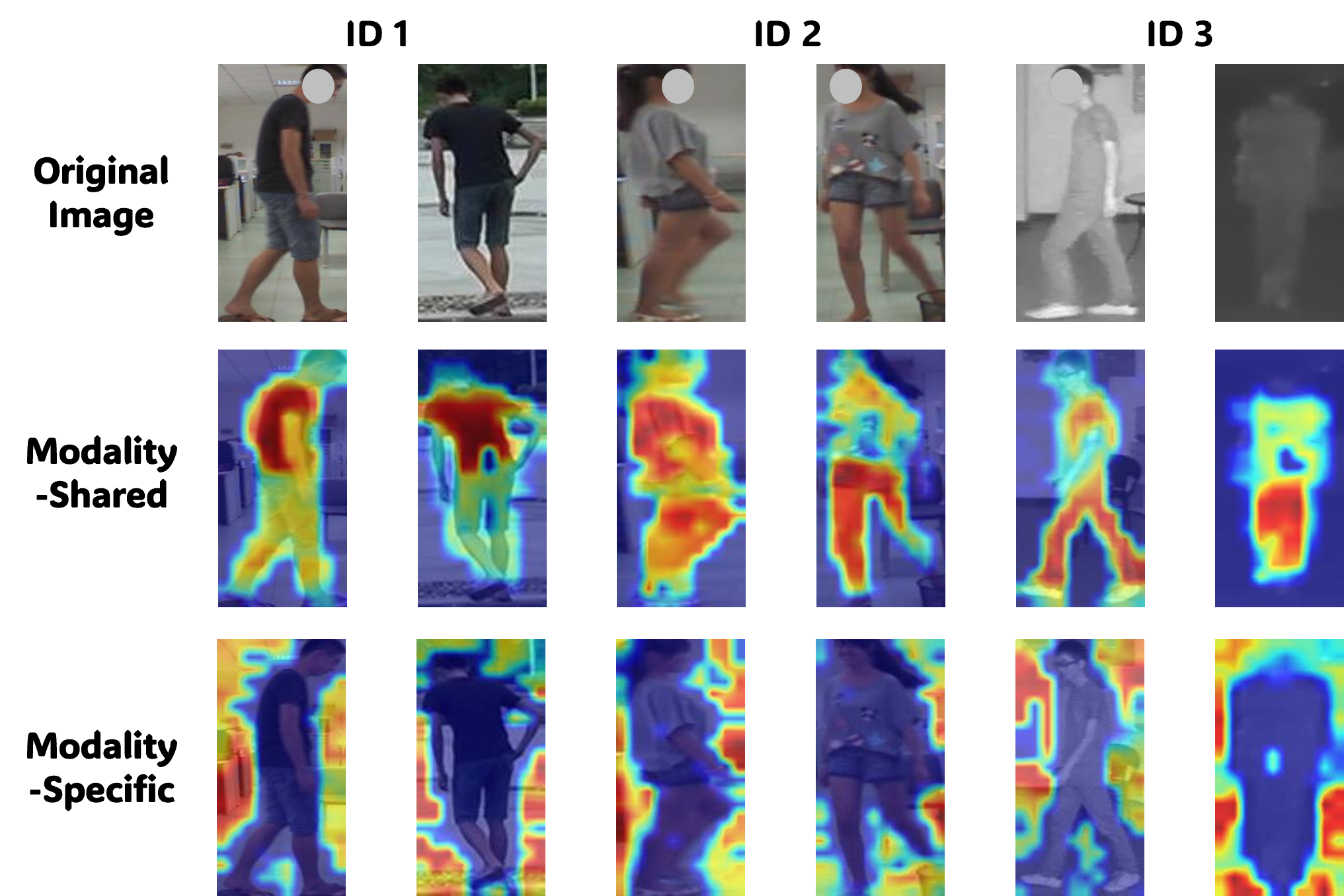}\vspace*{-5mm}
    \caption{Attention maps of modality-shared and modality-specific features for the same sample.}\vspace*{-4mm}
    \label{figfeature}
\end{figure}
Figure~\ref{figfeature} presents Grad-CAM~\cite{Selvaraju2017Grad} visualizations of modality-shared and modality-specific features extracted from the same image. As shown, modality-shared features mainly attend to modality-invariant information, such as body contours, which is identity-discriminative and consistent across modalities. In contrast, modality-specific features focus on modality-style information, such as color and other low-level imaging characteristics. The results show that modality-specific features, guided by $L_D$, do not merely duplicate modality-shared features but rather learn complementary modality-style information ignored in modality-shared features, thereby enhancing homogeneous-modality matching.

\vspace*{-2mm}
\section{Conclusion}

In this paper, we have investigated the limitations of conventional VI-ReID in open-world scenarios and introduced the Visible-Infrared Modality-Incomplete Re-Identification (VIMI-ReID) task. 
Unlike settings where the query and gallery are fixed to purely homogeneous or purely heterogeneous modalities, VIMI-ReID specifically addresses the challenge of modality incompleteness, where queries and galleries no longer adhere to fixed homogeneous or heterogeneous configurations, better reflecting the unpredictability of real-world data streams. To support research in this direction, we have reorganized the SYSU-MM01 and RegDB datasets into standardized VIMI-ReID benchmarks.
To overcome the trustworthiness challenges inherent to this new task, we proposed the Modality Adaptive Matching Transformer (MAMT). This framework synergistically leverages both modality-shared and modality-specific representations. The Shared Transformer Module (STM) extracts modality-invariant features to bridge the heterogeneous modality gap, while the Divergence Transformer Module (DTM) captures discriminative modality-style information that is often ignored in shared representations, thereby enhancing discrimination within the same modality. The Modality Adaptive Matching Module (MAM) dynamically fuses these complementary features based on the inferred query-gallery modality relationship, effectively mitigating matching conflicts, interference from modality uncertainty, and robustness degradation.
Our work underscores the critical need to address modality-incomplete retrieval for deploying robust person re-identification systems. We demonstrate that the intentional incorporation of modality-style information is crucial for enhancing the robustness and effectiveness of ReID in open-world conditions. Extensive experiments on the SYSU-VIMI and RegDB-VIMI benchmarks confirm that MAMT achieves state-of-the-art performance across all evaluation settings.

\bmhead{Data availability} The datasets used during and/or analyzed during the current study are available in the related public dataset repositories. The SYSU-MM01 dataset can be downloaded from https://github.com/wuancong/SYSU-MM01, the RegDB dataset can be downloaded from http://dm.dongguk.edu/link.html.

\bmhead{Acknowledgements} This work was supported by the Fundamental and Interdisciplinary Disciplines Breakthrough Plan of the Ministry of Education of China under Grant JYB2025XDXM906, the National Natural Science Foundation of China under Grant 62376201, the Hubei Provincial Science and Technology Talent Service Enterprise Project under Grant 2025DJB059, and the Hubei Provincial Special Fund for Central-Guided Local S\&T Development under Grant 2025CSA017.

\bibliography{sn-bibliography}

@inproceedings{Liu2024Cloth-aware,
author = {Liu, Fangyi and Ye, Mang and Du, Bo},
title = {Cloth-aware Augmentation for Cloth-generalized Person Re-identification},
year = {2024},
isbn = {9798400706868},
publisher = {Association for Computing Machinery},
address = {New York, NY, USA},
doi = {10.1145/3664647.3680956},
booktitle = {Proceedings of the 32nd ACM International Conference on Multimedia},
pages = {4053–4062},
numpages = {10},
location = {Melbourne VIC, Australia},
series = {MM '24}
}

@INPROCEEDINGS{Wu2017RGB,
  author={Wu, Ancong and Zheng, Wei-Shi and Yu, Hong-Xing and Gong, Shaogang and Lai, Jianhuang},
  booktitle={2017 IEEE International Conference on Computer Vision (ICCV)}, 
  title={RGB-Infrared Cross-Modality Person Re-identification}, 
  year={2017},
  volume={},
  number={},
  pages={5390-5399},
  doi={10.1109/ICCV.2017.575}}

@ARTICLE{Qiu2025Advancing,
  author={Qiu, Yuxuan and Wang, Liyang and Song, Wei and Liu, Jiawei and Shi, Zhiping and Jiang, Na},
  journal={IEEE Transactions on Information Forensics and Security}, 
  title={Advancing Visible-Infrared Person Re-Identification: Synergizing Visual-Textual Reasoning and Cross-Modal Feature Alignment}, 
  year={2025},
  volume={20},
  number={},
  pages={2184-2196},
  doi={10.1109/TIFS.2025.3539946}}

@inproceedings{Hu2024Empowering,
 author = {Hu, Zhangyi and Yang, Bin and Ye, Mang},
 booktitle = {Advances in Neural Information Processing Systems},
 doi = {10.52202/079017-3726},
 pages = {117363--117387},
 publisher = {Curran Associates, Inc.},
 title = {Empowering Visible-Infrared Person Re-Identification with Large Foundation Models},
 volume = {37},
 address = {Red Hook, NY, USA},
 year = {2024}
}

@ARTICLE{Pan2024Unified,
  author={Pan, Honghu and Pei, Wenjie and Li, Xin and He, Zhenyu},
  journal={IEEE Transactions on Information Forensics and Security}, 
  title={Unified Conditional Image Generation for Visible-Infrared Person Re-Identification}, 
  year={2024},
  volume={19},
  number={},
  pages={9026-9038},
  doi={10.1109/TIFS.2024.3426335}}

@ARTICLE{Cui2024DMA,
  author={Cui, Zhenyu and Zhou, Jiahuan and Peng, Yuxin},
  journal={IEEE Transactions on Information Forensics and Security}, 
  title={DMA: Dual Modality-Aware Alignment for Visible-Infrared Person Re-Identification}, 
  year={2024},
  volume={19},
  number={},
  pages={2696-2708},
  doi={10.1109/TIFS.2024.3352408}}

@ARTICLE{Sun2024Robust,
  author={Sun, Rui and Chen, Long and Zhang, Lei and Xie, Ruirui and Gao, Jun},
  journal={IEEE Transactions on Information Forensics and Security}, 
  title={Robust Visible-Infrared Person Re-Identification Based on Polymorphic Mask and Wavelet Graph Convolutional Network}, 
  year={2024},
  volume={19},
  number={},
  pages={2800-2813},
  doi={10.1109/TIFS.2024.3354377}}

@ARTICLE{Dou2023Hum,
  author={Dou, Shuguang and Zhao, Cairong and Jiang, Xinyang and Zhang, Shanshan and Zheng, Wei-Shi and Zuo, Wangmeng},
  journal={IEEE Transactions on Image Processing}, 
  title={Human Co-Parsing Guided Alignment for Occluded Person Re-Identification}, 
  year={2023},
  volume={32},
  number={},
  pages={458-470},
  doi={10.1109/TIP.2022.3229639}}

@inproceedings{ye2024dynamic,
title={Dynamic Feature Pruning and Consolidation for Occluded Person Re-identification}, 
volume={38}, 
DOI={10.1609/aaai.v38i7.28491}, 
abstractNote={Occluded person re-identification (ReID) is a challenging problem due to contamination from occluders. Existing approaches address the issue with prior knowledge cues, such as human body key points and semantic segmentations, which easily fail in the presence of heavy occlusion and other humans as occluders. In this paper, we propose a feature pruning and consolidation (FPC) framework to circumvent explicit human structure parsing. The framework mainly consists of a sparse encoder, a multi-view feature mathcing module, and a feature consolidation decoder. Specifically, the sparse encoder drops less important image tokens, mostly related to background noise and occluders, solely based on correlation within the class token attention. Subsequently, the matching stage relies on the preserved tokens produced by the sparse encoder to identify k-nearest neighbors in the gallery by measuring the image and patch-level combined similarity. Finally, we use the feature consolidation module to compensate pruned features using identified neighbors for recovering essential information while disregarding disturbance from noise and occlusion. Experimental results demonstrate the effectiveness of our proposed framework on occluded, partial, and holistic Re-ID datasets. In particular, our method outperforms state-of-the-art results by at least 8.6% mAP and 6.0% Rank-1 accuracy on the challenging Occluded-Duke dataset.}, 
number={7}, 
booktitle={Proceedings of the AAAI Conference on Artificial Intelligence}, 
author={Ye, YuTeng and Zhou, Hang and Cai, Jiale and Gao, Chenxing and Zhang, Youjia and Wang, Junle and Hu, Qiang and Yu, Junqing and Yang, Wei}, 
year={2024}, 
month={Mar.}, 
pages={6684-6692} }

@ARTICLE{Min2018Bli,
  author={Min, Xiongkuo and Zhai, Guangtao and Gu, Ke and Liu, Yutao and Yang, Xiaokang},
  journal={IEEE Transactions on Broadcasting}, 
  title={Blind Image Quality Estimation via Distortion Aggravation}, 
  year={2018},
  volume={64},
  number={2},
  pages={508-517},
  doi={10.1109/TBC.2018.2816783}}

@inproceedings{Huang2019Illumination,
author = {Huang, Yukun and Zha, Zheng-Jun and Fu, Xueyang and Zhang, Wei},
title = {Illumination-Invariant Person Re-Identification},
year = {2019},
isbn = {9781450368896},
publisher = {Association for Computing Machinery},
address = {New York, NY, USA},
doi = {10.1145/3343031.3350994},
booktitle = {Proceedings of the 27th ACM International Conference on Multimedia},
pages = {365–373},
numpages = {9},
location = {Nice, France},
series = {MM '19}
}

@ARTICLE{Lu2024Illumination,
  author={Lu, Andong and Zhang, Zhang and Huang, Yan and Zhang, Yifan and Li, Chenglong and Tang, Jin and Wang, Liang},
  journal={IEEE Transactions on Multimedia}, 
  title={Illumination Distillation Framework for Nighttime Person Re-Identification and a New Benchmark}, 
  year={2024},
  volume={26},
  number={},
  pages={406-419},
  doi={10.1109/TMM.2023.3266066}}

@article{Zhu2020Hetero,
title = {Hetero-Center loss for cross-modality person Re-identification},
journal = {Neurocomputing},
volume = {386},
pages = {97-109},
year = {2020},
issn = {0925-2312},
doi = {https://doi.org/10.1016/j.neucom.2019.12.100},
author = {Yuanxin Zhu and Zhao Yang and Li Wang and Sai Zhao and Xiao Hu and Dapeng Tao}
}

@ARTICLE{Liu2021Parameter,
  author={Liu, Haijun and Tan, Xiaoheng and Zhou, Xichuan},
  journal={IEEE Transactions on Multimedia}, 
  title={Parameter Sharing Exploration and Hetero-Center Triplet Loss for Visible-Thermal Person Re-Identification}, 
  year={2021},
  volume={23},
  number={},
  pages={4414-4425},
  doi={10.1109/TMM.2020.3042080}}

@inproceedings{Zhang2021Towards,
author = {Zhang, Yukang and Yan, Yan and Lu, Yang and Wang, Hanzi},
title = {Towards a Unified Middle Modality Learning for Visible-Infrared Person Re-Identification},
year = {2021},
isbn = {9781450386517},
publisher = {Association for Computing Machinery},
address = {New York, NY, USA},
doi = {10.1145/3474085.3475250},
booktitle = {Proceedings of the 29th ACM International Conference on Multimedia},
pages = {788–796},
numpages = {9},
location = {Virtual Event, China},
series = {MM '21}
}

@inproceedings{Qiu2024High,
title={High-Order Structure Based Middle-Feature Learning for Visible-Infrared Person Re-identification}, 
volume={38}, 
DOI={10.1609/aaai.v38i5.28259}, 
abstractNote={Visible-infrared person re-identification (VI-ReID) aims to retrieve images of the same persons captured by visible (VIS) and infrared (IR) cameras. Existing VI-ReID methods ignore high-order structure information of features while being relatively difficult to learn a reasonable common feature space due to the large modality discrepancy between VIS and IR images. To address the above problems, we propose a novel high-order structure based middle-feature learning network (HOS-Net) for effective VI-ReID. Specifically, we first leverage a short- and long-range feature extraction (SLE) module to effectively exploit both short-range and long-range features. Then, we propose a high-order structure learning (HSL) module to successfully model the high-order relationship across different local features of each person image based on a whitened hypergraph network. This greatly alleviates model collapse and enhances feature representations. Finally, we develop a common feature space learning (CFL) module to learn a discriminative and reasonable common feature space based on middle features generated by aligning features from different modalities and ranges. In particular, a modality-range identity-center contrastive (MRIC) loss is proposed to reduce the distances between the VIS, IR, and middle features, smoothing the training process. Extensive experiments on the SYSU-MM01, RegDB, and LLCM datasets show that our HOS-Net achieves superior state-of-the-art performance. Our code is available at https://github.com/Jaulaucoeng/HOS-Net.}, 
number={5}, 
booktitle={Proceedings of the AAAI Conference on Artificial Intelligence}, 
author={Qiu, Liuxiang and Chen, Si and Yan, Yan and Xue, Jing-Hao and Wang, Da-Han and Zhu, Shunzhi}, 
year={2024}, 
month={Mar.}, 
pages={4596-4604} }

@inproceedings{Zhang2023ProtoHPE,
author = {Zhang, Guiwei and Zhang, Yongfei and Tan, Zichang},
title = {ProtoHPE: Prototype-guided High-frequency Patch Enhancement for Visible-Infrared Person Re-identification},
year = {2023},
isbn = {9798400701085},
publisher = {Association for Computing Machinery},
address = {New York, NY, USA},
doi = {10.1145/3581783.3612297},
booktitle = {Proceedings of the 31st ACM International Conference on Multimedia},
pages = {944–954},
numpages = {11},
location = {Ottawa ON, Canada},
series = {MM '23}
}

@ARTICLE{Ye2024Channel,
  author={Ye, Mang and Wu, Zesen and Chen, Cuiqun and Du, Bo},
  journal={IEEE Transactions on Pattern Analysis and Machine Intelligence}, 
  title={Channel Augmentation for Visible-Infrared Re-Identification}, 
  year={2024},
  volume={46},
  number={4},
  pages={2299-2315},
  doi={10.1109/TPAMI.2023.3332875}}

@ARTICLE{Feng2020Learning,
  author={Feng, Zhanxiang and Lai, Jianhuang and Xie, Xiaohua},
  journal={IEEE Transactions on Image Processing}, 
  title={Learning Modality-Specific Representations for Visible-Infrared Person Re-Identification}, 
  year={2020},
  volume={29},
  number={},
  pages={579-590},
  doi={10.1109/TIP.2019.2928126}}

@INPROCEEDINGS{Lu2020Cross,
  author={Lu, Yan and Wu, Yue and Liu, Bin and Zhang, Tianzhu and Li, Baopu and Chu, Qi and Yu, Nenghai},
  booktitle={2020 IEEE/CVF Conference on Computer Vision and Pattern Recognition (CVPR)}, 
  title={Cross-Modality Person Re-Identification With Shared-Specific Feature Transfer}, 
  year={2020},
  volume={},
  number={},
  pages={13376-13386},
  doi={10.1109/CVPR42600.2020.01339}}

@INPROCEEDINGS{Zhang2022FMCNet,
  author={Zhang, Qiang and Lai, Changzhou and Liu, Jianan and Huang, Nianchang and Han, Jungong},
  booktitle={2022 IEEE/CVF Conference on Computer Vision and Pattern Recognition (CVPR)}, 
  title={FMCNet: Feature-Level Modality Compensation for Visible-Infrared Person Re-Identification}, 
  year={2022},
  volume={},
  number={},
  pages={7339-7348},
  doi={10.1109/CVPR52688.2022.00720}}

@ARTICLE{Liu2022Revisiting,
  author={Liu, Jianan and Wang, Jialiang and Huang, Nianchang and Zhang, Qiang and Han, Jungong},
  journal={IEEE Transactions on Circuits and Systems for Video Technology}, 
  title={Revisiting Modality-Specific Feature Compensation for Visible-Infrared Person Re-Identification}, 
  year={2022},
  volume={32},
  number={10},
  pages={7226-7240},
  doi={10.1109/TCSVT.2022.3168999}}

@ARTICLE{Zheng2022Visible,
  author={Zheng, Xiangtao and Chen, Xiumei and Lu, Xiaoqiang},
  journal={IEEE Transactions on Image Processing}, 
  title={Visible-Infrared Person Re-Identification via Partially Interactive Collaboration}, 
  year={2022},
  volume={31},
  number={},
  pages={6951-6963},
  doi={10.1109/TIP.2022.3217697}}

@ARTICLE{Li2022Visible,
  author={Li, Yulin and Zhang, Tianzhu and Liu, Xiang and Tian, Qi and Zhang, Yongdong and Wu, Feng},
  journal={IEEE Transactions on Image Processing}, 
  title={Visible-Infrared Person Re-Identification With Modality-Specific Memory Network}, 
  year={2022},
  volume={31},
  number={},
  pages={7165-7178},
  doi={10.1109/TIP.2022.3220408}}

@INPROCEEDINGS{Yu2023Modality,
  author={Yu, Hao and Cheng, Xu and Peng, Wei and Liu, Weihao and Zhao, Guoying},
  booktitle={2023 IEEE/CVF International Conference on Computer Vision (ICCV)}, 
  title={Modality Unifying Network for Visible-Infrared Person Re-Identification}, 
  year={2023},
  volume={},
  number={},
  pages={11151-11161},
  doi={10.1109/ICCV51070.2023.01027}}

@INPROCEEDINGS{Ren2024Implicit,
  author={Ren, Kaijie and Zhang, Lei},
  booktitle={2024 IEEE/CVF Conference on Computer Vision and Pattern Recognition (CVPR)}, 
  title={Implicit Discriminative Knowledge Learning for Visible-Infrared Person Re-Identification}, 
  year={2024},
  volume={},
  number={},
  pages={393-402},
  doi={10.1109/CVPR52733.2024.00045}}

@ARTICLE{Ye2022Deep,
  author={Ye, Mang and Shen, Jianbing and Lin, Gaojie and Xiang, Tao and Shao, Ling and Hoi, Steven C. H.},
  journal={IEEE Transactions on Pattern Analysis and Machine Intelligence}, 
  title={Deep Learning for Person Re-Identification: A Survey and Outlook}, 
  year={2022},
  volume={44},
  number={6},
  pages={2872-2893},
  doi={10.1109/TPAMI.2021.3054775}}

@Article{Nguyen2017Person,
AUTHOR = {Nguyen, Dat Tien and Hong, Hyung Gil and Kim, Ki Wan and Park, Kang Ryoung},
TITLE = {Person Recognition System Based on a Combination of Body Images from Visible Light and Thermal Cameras},
JOURNAL = {Sensors},
VOLUME = {17},
YEAR = {2017},
NUMBER = {3},
ARTICLE-NUMBER = {605},
PubMedID = {28300783},
ISSN = {1424-8220},
DOI = {10.3390/s17030605}
}

@INPROCEEDINGS{Zhang2023Diverse,
  author={Zhang, Yukang and Wang, Hanzi},
  booktitle={2023 IEEE/CVF Conference on Computer Vision and Pattern Recognition (CVPR)}, 
  title={Diverse Embedding Expansion Network and Low-Light Cross-Modality Benchmark for Visible-Infrared Person Re-identification}, 
  year={2023},
  volume={},
  number={},
  pages={2153-2162},
  doi={10.1109/CVPR52729.2023.00214}}

@inproceedings{Lu2023Learning,
title={Learning Progressive Modality-Shared Transformers for Effective Visible-Infrared Person Re-identification}, 
volume={37},  
DOI={10.1609/aaai.v37i2.25273}, 
abstractNote={Visible-Infrared Person Re-Identification (VI-ReID) is a challenging retrieval task under complex modality changes. Existing methods usually focus on extracting discriminative visual features while ignoring the reliability and commonality of visual features between different modalities. In this paper, we propose a novel deep learning framework named Progressive Modality-shared Transformer (PMT) for effective VI-ReID. To reduce the negative effect of modality gaps, we first take the gray-scale images as an auxiliary modality and propose a progressive learning strategy. Then, we propose a Modality-Shared Enhancement Loss (MSEL) to guide the model to explore more reliable identity information from modality-shared features. Finally, to cope with the problem of large intra-class differences and small inter-class differences, we propose a Discriminative Center Loss (DCL) combined with the MSEL to further improve the discrimination of reliable features. Extensive experiments on SYSU-MM01 and RegDB datasets show that our proposed framework performs better than most state-of-the-art methods. For model reproduction, we release the source code at https://github.com/hulu88/PMT.}, 
number={2}, 
booktitle={Proceedings of the AAAI Conference on Artificial Intelligence}, 
author={Lu, Hu and Zou, Xuezhang and Zhang, Pingping}, 
year={2023}, 
month={Jun.}, 
pages={1835-1843} }

@INPROCEEDINGS{Ye2021Channel,
  author={Ye, Mang and Ruan, Weijian and Du, Bo and Shou, Mike Zheng},
  booktitle={2021 IEEE/CVF International Conference on Computer Vision (ICCV)}, 
  title={Channel Augmented Joint Learning for Visible-Infrared Recognition}, 
  year={2021},
  volume={},
  number={},
  pages={13547-13556},
  doi={10.1109/ICCV48922.2021.01331}}

@INPROCEEDINGS{Selvaraju2017Grad,
  author={Selvaraju, Ramprasaath R. and Cogswell, Michael and Das, Abhishek and Vedantam, Ramakrishna and Parikh, Devi and Batra, Dhruv},
  booktitle={2017 IEEE International Conference on Computer Vision (ICCV)}, 
  title={Grad-CAM: Visual Explanations from Deep Networks via Gradient-Based Localization}, 
  year={2017},
  volume={},
  number={},
  pages={618-626},
  doi={10.1109/ICCV.2017.74}}

@INPROCEEDINGS{Feng2023Shape,
  author={Feng, Jiawei and Wu, Ancong and Zheng, Wei-Shi},
  booktitle={2023 IEEE/CVF Conference on Computer Vision and Pattern Recognition (CVPR)}, 
  title={Shape-Erased Feature Learning for Visible-Infrared Person Re-Identification}, 
  year={2023},
  volume={},
  number={},
  pages={22752-22761},
  doi={10.1109/CVPR52729.2023.02179}}

@article{zhang2025adaptive,
author={Zhang, Yukang
and Yan, Yan
and Lu, Yang
and Wang, Hanzi},
title={Adaptive Middle Modality Alignment Learning for Visible-Infrared Person Re-identification},
journal={International Journal of Computer Vision},
year={2025},
month={Apr},
day={01},
volume={133},
number={4},
pages={2176-2196},
issn={1573-1405},
doi={10.1007/s11263-024-02276-4}
}

@InProceedings{He_2024_CVPR,
  author={He, Weizhen and Deng, Yiheng and Tang, Shixiang and Chen, Qihao and Xie, Qingsong and Wang, Yizhou and Bai, Lei and Zhu, Feng and Zhao, Rui and Ouyang, Wanli and Qi, Donglian and Yan, Yunfeng},
  booktitle={2024 IEEE/CVF Conference on Computer Vision and Pattern Recognition (CVPR)}, 
  title={Instruct-ReID: A Multi-Purpose Person Re-Identification Task with Instructions}, 
  year={2024},
  volume={},
  number={},
  pages={17521-17531},
  doi={10.1109/CVPR52733.2024.01659}}

@Article{Zhang2025Modality,
author={Zhang, Yiyuan
and Zhao, Sanyuan
and Ye, Mang
and Yang, Ruigang
and Shen, Jianbing},
title={Modality Confusion Learning: A Versatile Framework for Visible-Infrared Re-identification},
journal={International Journal of Computer Vision},
year={2025},
month={Dec},
day={01},
volume={133},
number={12},
pages={8469-8488},
issn={1573-1405},
doi={10.1007/s11263-025-02563-8}
}

@Article{Wu2020RGB-IR,
author={Wu, Ancong
and Zheng, Wei-Shi
and Gong, Shaogang
and Lai, Jianhuang},
title={RGB-IR Person Re-identification by Cross-Modality Similarity Preservation},
journal={International Journal of Computer Vision},
year={2020},
month={Jun},
day={01},
volume={128},
number={6},
pages={1765-1785},
issn={1573-1405},
doi={10.1007/s11263-019-01290-1}
}

@Article{Yin2020Fine,
author={Yin, Jiahang
and Wu, Ancong
and Zheng, Wei-Shi},
title={Fine-Grained Person Re-identification},
journal={International Journal of Computer Vision},
year={2020},
month={Jun},
day={01},
volume={128},
number={6},
pages={1654-1672},
issn={1573-1405},
doi={10.1007/s11263-019-01259-0}
}

@Article{Ye2025Transformer,
author={Ye, Mang
and Chen, Shuoyi
and Li, Chenyue
and Zheng, Wei-Shi
and Crandall, David
and Du, Bo},
title={Transformer for Object Re-identification: A Survey},
journal={International Journal of Computer Vision},
year={2025},
month={May},
day={01},
volume={133},
number={5},
pages={2410-2440},
issn={1573-1405},
doi={10.1007/s11263-024-02284-4}
}

@ARTICLE{Zhong2022Grayscale,
  author={Zhong, Xian and Lu, Tianyou and Huang, Wenxin and Ye, Mang and Jia, Xuemei and Lin, Chia-Wen},
  journal={IEEE Transactions on Circuits and Systems for Video Technology}, 
  title={Grayscale Enhancement Colorization Network for Visible-Infrared Person Re-Identification}, 
  year={2022},
  volume={32},
  number={3},
  pages={1418-1430},
  doi={10.1109/TCSVT.2021.3072171}}

@inproceedings{dai2025diffusion,
title={Diffusion-based Synthetic Data Generation for Visible-Infrared Person Re-Identification}, 
volume={39}, 
DOI={10.1609/aaai.v39i11.33216}, 
abstractNote={The performance of models is intricately linked to the abundance of training data. In Visible-Infrared person Re-IDentification (VI-ReID) tasks, collecting and annotating large-scale images of each individual under various cameras and modalities is tedious, time-expensive, costly and must comply with data protection laws, posing a severe challenge in meeting dataset requirements. Current research investigates the generation of synthetic data as an efficient and privacy-ensuring alternative to collecting real data in the field. However, a specific data synthesis technique tailored for VI-ReID models has yet to be explored. In this paper, we present a novel data generation framework, dubbed Diffusion-based VI-ReID data Expansion (DiVE), that automatically obtain massive RGB-IR paired images with identity preserving by decoupling identity and modality to improve the performance of VI-ReID models. Specifically, identity representation is acquired from a set of samples sharing the same ID, whereas the modality of images is learned by fine-tuning the Stable Diffusion (SD) on modality-specific data. DiVE extend the text-driven image synthesis to identity-preserving RGB-IR multimodal image synthesis. This approach significantly reduces data collection and annotation costs by directly incorporating synthetic data into ReID model training. Experiments have demonstrated that VI-ReID models trained on synthetic data produced by DiVE consistently exhibit notable enhancements. In particular, the state-of-the-art method, CAJ, trained with synthetic images, achieves an improvement of about 9% in mAP over the baseline on the LLCM dataset.}, 
number={11}, 
booktitle={Proceedings of the AAAI Conference on Artificial Intelligence}, 
author={Dai, Wenbo and Lu, Lijing and Li, Zhihang}, 
year={2025}, 
month={Apr.}, 
pages={11185-11193} }

@inproceedings{Wang2025Low,
author = {Wang, Dengwen and Xing, Guanyu and Liu, Yanli},
title = {Low-light Invariant Representation Learning for Visible-Infrared Person Re-identification},
year = {2025},
isbn = {9798400720352},
publisher = {Association for Computing Machinery},
address = {New York, NY, USA},
doi = {10.1145/3746027.3755601},
booktitle = {Proceedings of the 33rd ACM International Conference on Multimedia},
pages = {8645–8653},
numpages = {9},
location = {Dublin, Ireland},
series = {MM '25}
}

@inproceedings{Dosovitskiy2021Image,
  title={An Image is Worth 16x16 Words: Transformers for Image Recognition at Scale},
  author={Dosovitskiy, Alexey and Beyer, Lucas and Kolesnikov, Alexander and Weissenborn, Dirk and Zhai, Xiaohua and Unterthiner, Thomas and Dehghani, Mostafa and Minderer, Matthias and Heigold, Georg and Gelly, Sylvain and Uszkoreit, Jakob and Houlsby, Neil},
  booktitle={International Conference on Learning Representations (ICLR)},
  year={2021}
}

@INPROCEEDINGS{Deng2009ImageNet,
  author={Deng, Jia and Dong, Wei and Socher, Richard and Li, Li-Jia and Kai Li and Li Fei-Fei},
  booktitle={2009 IEEE Conference on Computer Vision and Pattern Recognition}, 
  title={ImageNet: A large-scale hierarchical image database}, 
  year={2009},
  volume={},
  number={},
  pages={248-255},
  doi={10.1109/CVPR.2009.5206848}}

\end{document}